\documentclass[runningheads]{llncs}

\usepackage{eccv}

\usepackage{eccvabbrv}

\usepackage{graphicx}
\usepackage{booktabs}
\usepackage{graphicx}  
\usepackage{tabularray}
\usepackage{array}
\usepackage{float}
\usepackage{ragg ed2e}
\usepackage{booktabs}
\usepackage{colortbl}
\usepackage{amsmath}
\usepackage{url}
\usepackage{utfsym}
\usepackage{graphicx}
\usepackage{makecell, multirow}
\usepackage{adjustbox}
\usepackage{xspace}
\usepackage{booktabs}
\usepackage{color}
\usepackage{amssymb,mathrsfs,amsmath}
\usepackage{arydshln} 
\usepackage{amsfonts} 
\definecolor{hollywoodcerise}{rgb}{0.96, 0.0, 0.63}
\definecolor{lasallegreen}{rgb}{0.03, 0.47, 0.19}
\definecolor{hanpurple}{rgb}{0.32, 0.09, 0.98}
\definecolor{green(pigment)}{rgb}{0.0, 0.65, 0.31}
\usepackage{bbding} 
\usepackage{hyperref}
\hypersetup{colorlinks,linkcolor={red},citecolor={hollywoodcerise},urlcolor={red}}  
\usepackage[accsupp]{axessibility}  

\usepackage{hyperref}

\usepackage{orcidlink}

\begin{document}

\title{EventBind: Learning a Unified Representation to Bind Them All for Event-based Open-world Understanding} 

\titlerunning{EventBind}

\author{Jiazhou Zhou\inst{1}\orcidlink{0009-0008-5258-1675} \and
Xu Zheng\inst{1}\orcidlink{0000-0003-4008-8951} \and
Yuanhuiyi Lyu\inst{1}\orcidlink{0009-0004-1450-811X} \and
Lin Wang\inst{1,2}\orcidlink{0000-0002-7485-4493}\thanks{Corresponding author}}

\authorrunning{J. Zhou et al.}

\institute{Hong Kong University of Science and Technology, Guangzhou, China \email{\{jiazhouzhou,yuanhuiyilv\}@hkust-gz.edu.cn, zhengxu128@gmail.com}
\and
Hong Kong University of Science and Technology, Hong Kong, China \\
\email{linwang@ust.hk} \\
\small{Project Page: \url{https://vlislab22.github.io/EventBind/}}}
\maketitle

\begin{abstract}
In this paper, we propose EventBind, a novel and effective framework that unleashes the potential of vision-language models (VLMs) for event-based recognition to compensate for the lack of large-scale event-based datasets. 
In particular, due to the distinct modality gap with the image-text data and the lack of large-scale datasets, learning a common representation space for images, texts, and events is non-trivial.  
Intuitively, we need to address two key challenges: 1) how to generalize CLIP's visual encoder to event data while fully leveraging events' unique properties, \eg, sparsity and high temporal 
resolution; 2) how to effectively align the multi-modal embeddings, \ie, image, text, and events. 
Accordingly, we first introduce a novel event encoder that subtly models the temporal information from events and meanwhile generates event prompts for modality bridging. We then design a text encoder that generates content prompts and utilizes hybrid text prompts to enhance EventBind's generalization ability across diverse datasets.
With the proposed event encoder, text encoder, and image encoder, a novel Hierarchical Triple Contrastive Alignment HTCA module is introduced to jointly optimize the correlation and enable efficient knowledge transfer among the three modalities. We evaluate various settings, including fine-tuning and few-shot on three benchmarks and our EventBind achieves new state-of-art accuracy compared with the previous methods, such as on N-Caltech101 (+5.34\% and +1.70\%) and N-Imagenet (+5.65\% and +1.99\%) with fine-tuning and 20-shot settings respectively. Moreover, our EventBind can be flexibly extended to the event retrieval task using text or image queries, showing plausible performance. 
\end{abstract}

\section{Introduction}\label{sec:introduction}

\begin{figure}[t!]
\centering
\includegraphics[width=\linewidth]{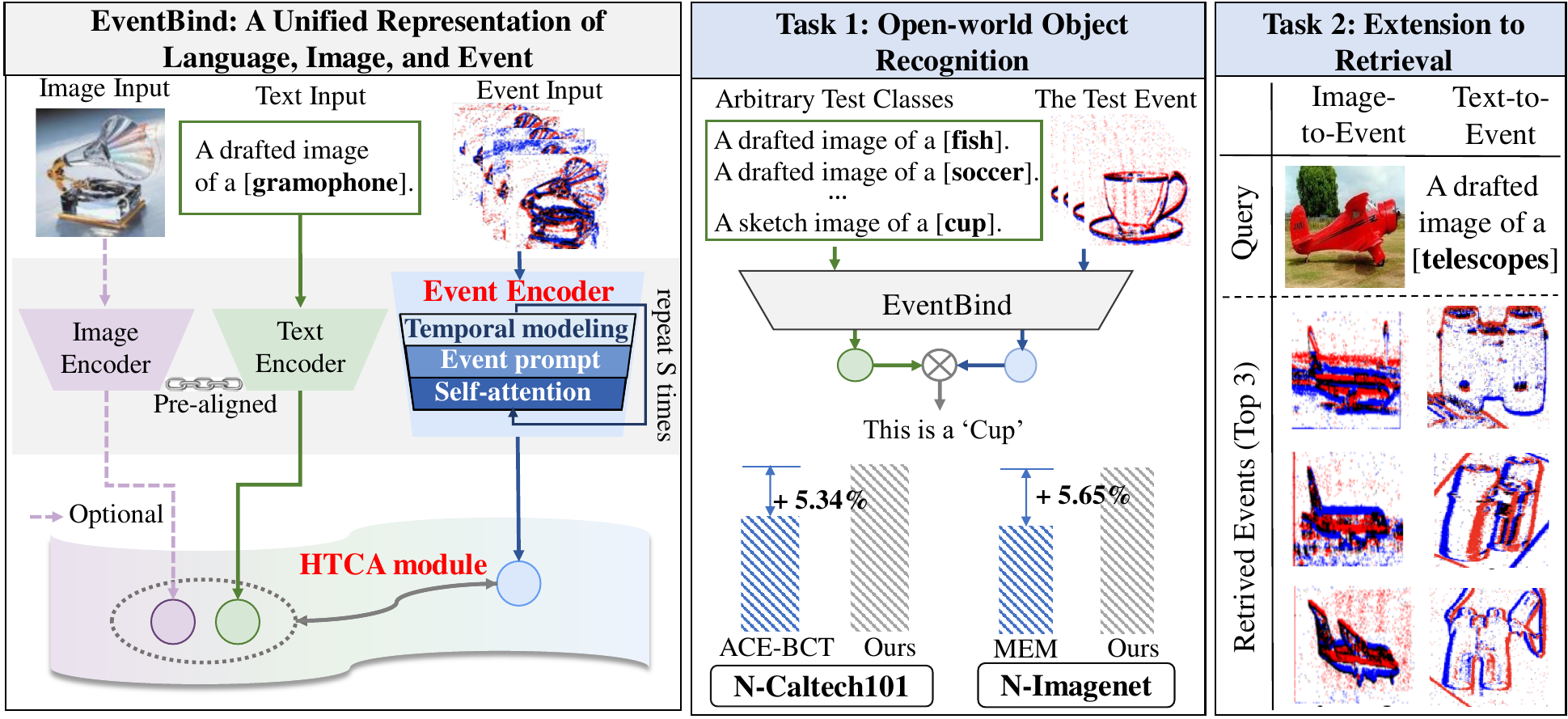}
\caption{\textbf{Overview of our EventBind}, which extracts the events' high temporal and sparse spatial information via the proposed Event Encoder and aligns event, image, and text embeddings in the unified representation space with a novel Hierarchical Triple Contrastive Alignment (HTCA) module. EventBind solves various practical tasks like open-world object recognition and few-shot object recognition with significant performance improvements compared to the previous best models~\cite{liu2022fast,klenk2024masked}. Our EventBind framework can be flexibly extended to image-to-event and text-to-event retrieval tasks.}\label{TesearFig}
\end{figure}

Event cameras are bio-inspired sensors~\cite{zheng2023deep,gallego2020event} that have recently received much attention in the computer vision and robotics community for their distinct merits, such as high temporal resolution and no motion blur.
Event cameras perceive the per-pixel brightness changes asynchronously and output event streams, encoding the time, pixel location, and polarity of intensity changes.
This distinct feature has sparked many research endeavors targeted at event cameras, and recently deep neural networks (DNNs) have been applied to event-based vision, showing significant performance gains for many tasks, such as object recognition~\cite{gehrig2019end, kim2021n, wang2019ev, Hots, gu2020tactilesgnet, li2021graph, liu2022fast}.
Event cameras pose superior performance in capturing objects in dynamic environments; however, learning high-performance DNNs for event data is often impeded by the asynchronous nature of events and the challenge of obtaining high-quality and large-scale labeled datasets~\cite{zheng2023deep, wu2023eventclip, gallego2020event}. Moreover, in real-world scenarios, DNN model failures may occur when encountering event data with new categories not present in the training set. Nonetheless, retraining large models for each new category is impractical, making it necessary to explore zero-shot and few-shot recognition pipelines for event cameras.

Recently, vision-language models (VLMs), \eg, CLIP~\cite{CLIP}, have shown promising open-world performance on the 2D image-based recognition tasks. 
Benefiting from large-scale training data (over 400M image-text pairs), CLIP can serve as the pre-trained model and be transferred to other visual data, \eg, video~\cite{wasim2023vita,ni2022expanding, rasheed2023fine} and depth~\cite{zhang2022can}, under the few-shot setting.
Intuitively, we ask a question: how to transfer the pre-trained CLIP to the event data and achieve open-world few-shot recognition performance while considering its asynchronous feature and the distinct modality shift from the image and text data?
To this end, we strive to address two crucial challenges. 
\textbf{1)} how to generalize CLIP's visual encoder to the event data while fully leveraging events' unique properties, \eg, sparsity and high temporal resolution. That is, the distinct modality discrepancy of event data, compared with the canonical images, makes it difficult to directly extract the spatial-temporal features if using CLIP's visual encoders.
\textbf{2)} how to effectively align the multi-modal embeddings, \ie, image, text, and event. The significant modality gap between them poses obstacles to reliable and effective feature alignments.

In this paper, we propose EventBind, a novel framework that unleashes the potential of CLIP for event-based recognition tasks to compensate for the lack of large-scale event-based datasets. 
Our EventBind consists of an event encoder, a text encoder, and the CLIP image encoder, as illustrated in Fig.~\ref{TesearFig}. 
Our method enjoys three key technical breakthroughs.
Firstly, we introduce an event encoder to address the challenge of generalizing CLIP's original visual encoder to the event data (Sec.~\ref{Section: Domain-gap Shifted Event Encoder}).
The event encoder incorporates event temporal modeling and event prompts generation to better exploit events' unique properties, such as sparsity and high temporal resolution. 
Specifically, the event temporal modeling enables temporal information exchange between event frames, while the generated event prompts are used to capture the spatial-temporal information of raw events.
To better align text with events, we introduce a new text encoder in our EventBind, building upon the basic text encoder in CLIP (Sec.~\ref{Section: Text Encoder}). The text encoder generates content prompts to improve the fine-tuning performance with a lightweight MLP network and incorporates hybrid text prompts~\textemdash combining hand-crafted and learnable prompts~\textemdash to enhance the generalization across diverse datasets. 
Moreover, we introduce an additional loss function to ensure consistency between the hand-crafted and learnable prompts. 

To tackle the second challenge, we propose a Hierarchical Triple Contrastive Alignment (HTCA) module to align the multi-modal feature embeddings, \ie, events, image, and text to learn a unified feature representation (Sec.~\ref{section: Hierarchical Triple Alignment}).
Concretely, multi-modal triple feature alignment is conducted to jointly align the text, events, and image by minimizing the contrastive loss between each two of them. Meanwhile, it imposes the semantic feature alignment to keep the semantic consistency between the events and image.
In a nutshell, with the HTCA module, we can effectively bridge the modality gap and facilitate efficient knowledge transfer among the three modalities.

We conduct extensive experiments to evaluate our EventBind on three event-based recognition benchmarks: N-Caltech101, N-MNIST, and N-ImageNet, covering fine-tuning and few-shot settings. The experimental results demonstrate that our EventBind significantly outperforms the existing methods by a large margin (Tab.~\ref{Fine-turned performance} and Tab.~\ref{Zero-shot table}) and the proposed event encoder shows superior efficiency (Tab.~\ref{ab_model_parameter}), thus enhancing downstream tasks in event-based vision. Additionally, we demonstrate that our EventBind can be flexibly extended to the application of the event retrieval tasks by utilizing both text and image queries, which further spotlights EventBind's transferability and versatility.

In summary, our main contributions are as follows: \textbf{(I)} We propose EventBind, a novel framework that unleashes the potential of CLIP for event-based recognition to compensate for the lack of large-scale datasets. (\textbf{II}) We propose a novel event encoder and a text encoder to harness properties of events (\eg, high temporal resolution) and to enhance the EventBind's generalization ability, respectively. (\textbf{III}) We propose a novel Hierarchical Triple Contrastive Alignment (HTCA) module that jointly optimizes the correlation alignment among three modalities. (\textbf{IV}) We demonstrate that our EventBind outperforms existing methods by a significant margin in both fine-tuning and few-shot settings on three event recognition benchmarks. As a penitential, our EventBind can also be freely extended to the event retrieval task when employing text or image queries.
    
\section{Related Works}\label{sec:Related Works}

\noindent\textbf{Vision-language Models (VLMs).}
VLMs gain interest for their cross-modal transfer ability, focusing on aligning image-text embeddings rather than just image data. As a pioneering work, CLIP~\cite{CLIP} employs contrastive learning objectives on image-text pairs and achieves impressive zero-shot classification on 30+ datasets. Motivated by this, other works have aligned the original image and text with different types of modalities, such as audio~\cite{mahmud2023ave}, 3D point clouds~\cite{zhang2022pointclip, zeng2023clip2, chen2023vlp, xue2023ulip}, video~\cite{wasim2023vita,ni2022expanding, rasheed2023fine}, and depth~\cite{zhang2022can}. In short, methods can be split into three types: 1) Single-encoder framework with frozen CLIP visual encoder as backbone~\cite{huang2022frozen, lin2022frozen}; 2) Dual-encoder framework using CLIP's text and fine-tuned visual encoder for added modality~\cite{zhang2022pointclip, liu2023partslip,wasim2023vita,ni2022expanding, zhang2022can, rasheed2023fine}; 3) Triple-encoder framework with CLIP's text and visual encoders and new encoder for additional modality embeddings~\cite{mahmud2023ave,zeng2023clip2, xue2023ulip}. However, there are few attempts to align event modality with image and text. One concurrent work called EventCLIP~\cite{wu2023eventclip} exploits the gray-scale event frame as the intermediate representation of event data and utilizes the CLIP visual encoder for event embeddings. Our EventBind differs from ~\cite{wu2023eventclip} in three main aspects: 1) EventBind proposes a tailored event encoder instead of CLIP's image encoder; 2) EventBind creates a unified representation space for event, image, and text, unlike EventCLIP which only aligns event with text; 3) EventBind shows effectiveness in event retrieval alongside object recognition.

\noindent \textbf{Prompt Learning for VLMs.}
Prompt learning plays a pivotal role in harnessing the power of large-scale VLMs~\cite{chen2023vlp, du2022survey}. Existing research focuses on creating prompts for VLMs that consider both vision and language. Text prompts activate specific capabilities of VLMs through designed templates or learnable embeddings.~\cite{CLIP, zhou2022conditional, zhou2022learning, yao2023visual}. These prompts guide VLMs in cross-modal understanding, yielding promising results on various tasks~\cite{tsimpoukelli2021multimodal, wang2021simvlm, shen2022multitask, li2022align, li2023blip}. Visual prompts tuning injects a small number of learnable parameters into input space~\cite{jia2022visual}, enhancing the performance of VLMs~\cite{bahng2022exploring, bar2022visual, zang2022unified} and enabling rapid adaptation to downstream tasks~\cite{herzig2022promptonomyvit,bai2023textir}. Compared to text prompts, visual prompts can be flexibly presented in various forms, such as bounding boxes~\cite{bahng2022exploring,li2020oscar}, colored blocks~\cite{yao2021cpt}, positions~\cite{wang2023position} and points~\cite{kirillov2023segment}. In EventBind, we create event prompts to help the model learn event data features.

\noindent\textbf{Event-based Recognition.}
Event cameras excel in high-temporal resolution, low latency, and very high dynamic range, making them ideal for real-time object recognition in applications like autonomous vehicles and mobile systems~\cite{zheng2023deep}. However, event data's unique traits hinder the direct use of pre-trained models. For this reason, various methods~\cite{gallego2020event, kim2021n, zheng2023deep, wang2019ev, sironi2018hats, IETS, EST, gu2020tactilesgnet, botzheim2012human, amir2017low, Ev-VGCNN, li2021graph, liu2022fast} have been proposed.
However, learning high-performance DNNs for event data is challenging due to the asynchronous nature of events and the lack of large-scale labeled datasets~\cite{zheng2023deep}. Besides, DNNs may fail with new event categories, and retraining large models for each new category is impractical, so few-shot recognition pipelines for event cameras need to be explored. Intuitively, we propose EventBind with novel event encoders and specialized text prompts, buttressed by the HTCT module for effective event-based recognition.

\section{Proposed Approach}\label{sec:Proposed Method}
\begin{figure}[t!] 
	\centering
	\includegraphics[width=1.0\textwidth]{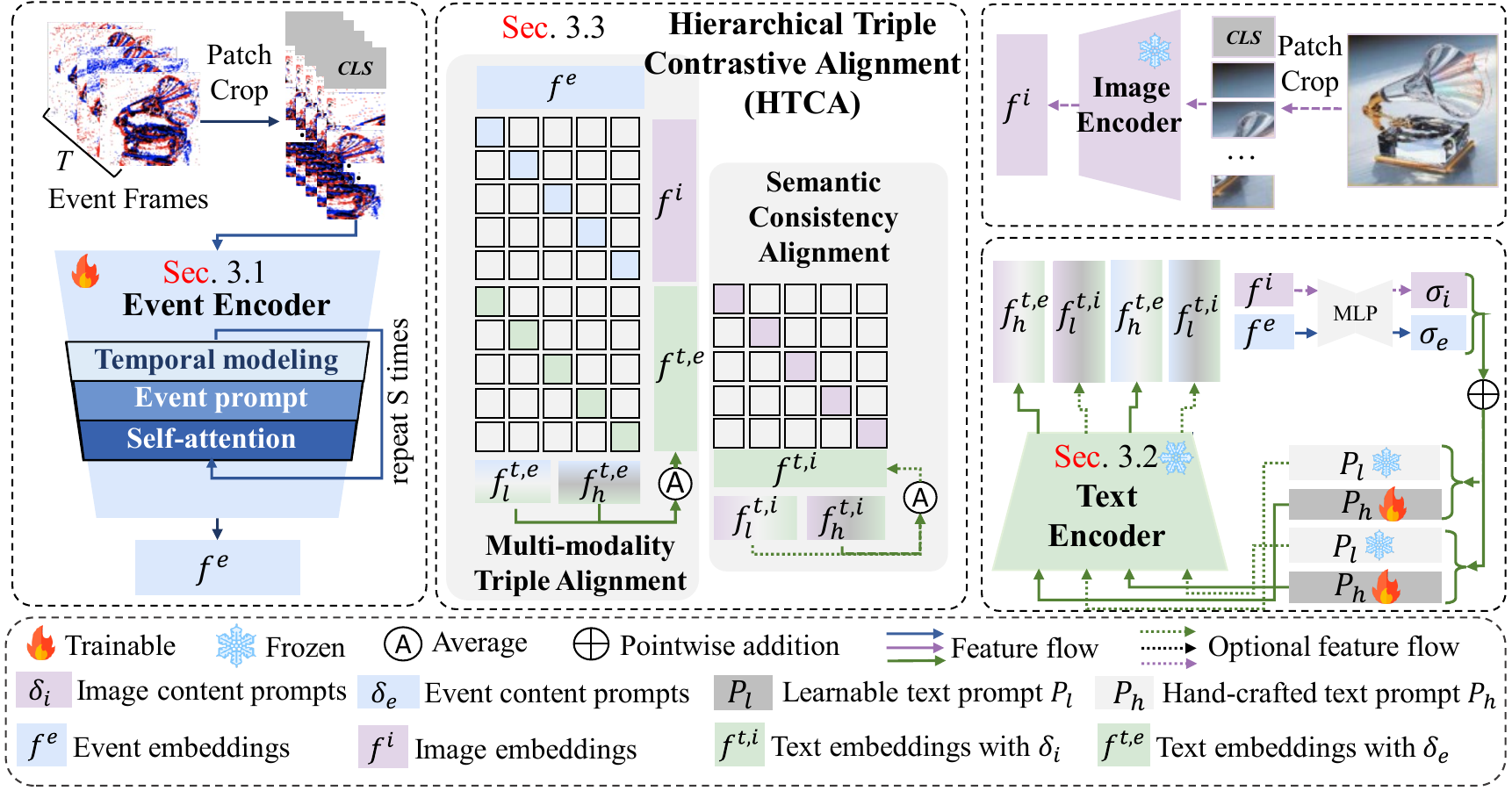}
	\caption{\textbf{Overview framework of EventBind}, which consists of image encoder pre-aligned with text encoder and the proposed event encoder. It takes image(optional), text, and event as input, generating the image embeddings $f^{i}$, event embeddings $f^{e}$ and text embeddings $f^{t,e}, f^{t,i}$. Then all output embeddings are aligned in the HTCA module to establish a unified representation space.}
 \label{pic:framework}
\end{figure}%

The proposed EventBind framework is presented in Fig.~\ref{pic:framework}, aiming at unleashing the potential of CLIP for event-based recognition tasks to compensate for the scarcity of large-scale event-based datasets. To this end, our EventBind addresses two technical challenges: 1) how to generalize the CLIP's visual encoder to the event data while fully exploiting the event's unique properties, \eg, spatial sparsity and high temporal resolution; 2) how to effectively align the multi-modal embeddings, \ie, image, text, and event, to establish a unified representation space while preserving CLIP's exceptional few-shot capability. 


Consequently, we design our EventBind consisting of four major components, as depicted in Fig.~\ref{pic:framework}: 
1) the event encoder, responsible for extracting dense temporal and sparse spatial information from events and bridging the modality gap (Sec.~\ref{Section: Domain-gap Shifted Event Encoder}); 2) the text encoder (Sec.~\ref{Section: Text Encoder}) incorporates the event and image content and pre-trained generalizing knowledge from CLIP to generate the text embeddings for enhance performance; 
3) the original CLIP's image encoder for semantically aligned image embeddings
\footnote{As the image encoder in EventBind remains identical to CLIP, we do not elaborate on it in the following subsection. Note that the image input is optional.};
4) The hierarchical triple contrastive alignment (HTCA) module (Sec.~\ref{section: Hierarchical Triple Alignment}) jointly optimizes the correlation among the three distinct modalities, forming a unified embedding space among image, text, and event. In the following sections, we describe these technical components.
\subsection{Event Encoder}
\label{Section: Domain-gap Shifted Event Encoder}
Event Encoder is designed to transfer the remarkable capability of CLIP to the asynchronous event-based vision. However, addressing the modality disparity among events, images, and text data is nontrivial. In this regard, it is crucial to design a tailored event encoder, which can fully leverage events' unique properties, such as high temporal information and sparsity. We propose a novel event encoder, that consists of two key technical parts: 1) event temporal modeling to effectively exploit the unique temporal information among event frames and 2) event prompts generation as additional modality-bridging parameters without introducing excessive parameters, thus enabling its efficiency and transferability.

\begin{figure}[t!]
\centering
\includegraphics[width=1.0\columnwidth]{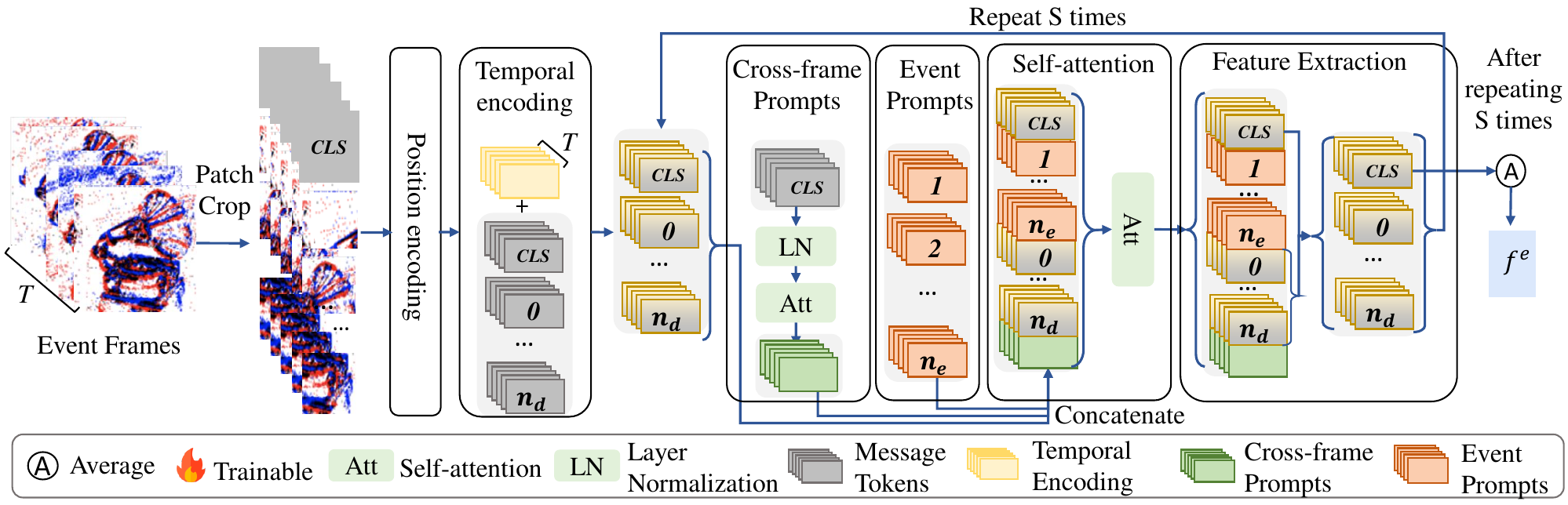}
\caption{The architecture of our event encoder consists of two key technical parts: (a) Temporal modeling consisting of temporal encoding and cross-frame prompts for event spatial-temporal modeling by introducing information exchange between event frames; (b) Event prompts generate event modality prompts to provide additional parameters for modality bridging.} \label{Event Prompts Generation Branch}
\end{figure}

Specifically, we transform the raw event into a series of event frames $I_{e} \in \mathbb{R}^{T \times H \times W \times 3}$ with $T$ frames of spatial size $H \times W$ (\textit{See more details of event frame representation in the suppl.}). Then each frame is divided into $L$ non-overlapping $P \times P$ square patches, where $L = H \times W / P^{2}$. For each event frame, all patches are flattened into a $3P^{2}$-dimension vector and presented as $\{x_{t, i} \in \mathbb{R}^{3P^{2}}\}^{L}_{i}$, where $t$ denotes the frame number and $i$ represents the patch number. All patch vectors are projected to $D$ dimension by the matrix multiplication with the linear projection $P_{emb} \in \mathbb{R}^{3P^{2} \times D }$. Besides, the additional class token, namely $CLS$, is concatenated at the beginning of the patch token sequence for each frame. The input token sequence $z_{t} \in \mathbb{R}^{(L+1) \times D}$ of event frame $t$ can be formulated as:
\begin{equation}
\setlength{\abovedisplayskip}{3pt}
\setlength{\belowdisplayskip}{3pt}
z_{t} = LN( [CLS, P_{emb}^{T}x_{t,1}, P_{emb}^{T} x_{t,2}, ..., P_{emb}^{T}x_{t,L}]  ),
\end{equation}
where $LN(.)$ represents the layer normalization, $[.]$ means the concatenation operation; $x_{t,L}$ denotes the original message tokens, where $t$ is the event frame number and $L$ is the total number of patches.

\noindent\textbf{Event Temporal Modeling.}
To model the unique temporal correlation among event frames, we add the temporal encoding $e$ with the input token sequence $z_{t}$ of the event encoder, where $e = e_{spatial} + e_{temporal}$. $e_{spatial}$ and $e_{temporal}$ denote the spatial and temporal positional encoding, respectively.
Following CLIP~\cite{CLIP}, $e_{spatial}$ and $e_{temporal}$ are learnable parameters and randomly initialized. Despite the temporal encoding, the event encoder also introduces cross-frame prompts to exchange information among event frames for better temporal information transmission. 
Concretely, shown in Fig.~\ref{Event Prompts Generation Branch}, for the $s$-th layer of the event encoder, 
the class tokens among all $T$ event frames are extracted from the input sequence $z_{t}^{s}, t=1,2,...,T$ and then fed into a norm layer~\cite{ba2016layer} and self-attention\cite{vaswani2017attention} successively, aiming at obtaining the $s$-\textit{th} layer cross frame prompts $x_{cf}^{s, t}$ as follows:
\begin{equation}
\setlength{\abovedisplayskip}{3pt}
\setlength{\belowdisplayskip}{3pt}
x_{cf}^{s,t} = \text{Att}(\text{LN}([x_{cf}^{s,1}, x_{cf}^{s,2}, ..., x_{cf}^{s,T}])),
\end{equation}
where $[.]$ is the concatenation operation. As shown in Fig.~\ref{Event Prompts Generation Branch} green rectangles, the obtained cross-frame prompts, which perceive the temporal relation among frames, are concatenated with the original token sequence $z_{t}^{s}$ as the input of the next $(s+1)$-\textit{th} layer of the event encoder. \\
\noindent\textbf{Event Prompts Generation.} 
Based on the event temporal modeling, the unique temporal correlation among event frames is captured. Since fine-tuning the pre-trained image encoder for processing event data may not guarantee good performance because of the modality gap and smaller event dataset size compared to the web-scale image dataset, we propose the event prompts to bridge the modality gaps among the image, text, and event modalities without introducing excessive parameters, which may cause catastrophic forgetting~\cite{jia2022visual}. Specifically, we insert a small number of learnable parameters at the input of each encoder layer. The proposed event prompts $P_{e}$ are learnable vectors inserted at the output token sequence between the original class token and other message tokens (see orange rectangles in Fig.~\ref{Event Prompts Generation Branch}). 
After event temporal modeling and event prompts generation, 
we concatenate the original message token $Z^{s,t}$, event prompts $P^{s,t}_{e}$ and cross-frame prompts $x_{cf}^{s,t}$ then send it into a standard self-attention layer as:
\begin{equation}
\setlength{\abovedisplayskip}{3pt}
\setlength{\belowdisplayskip}{3pt}
z_{t}^{s+1} = \text{Att}([CLS, P^{s,t}_{e}, Z^{s,t}, x_{cf}^{s,t}]),
\end{equation}
where $P^{s,t}_{e}$ are the event modality prompts for the $t$\textit{-th} frame in the $s$\textit{-th} layer and $P^{s,t}_{e} = [p_{1}^{s,t},p_{2}^{l,t},..., p_{n_{e}}^{s,t}]$; $n_{e}$ is the number of event modality prompts; $Z^{s,t}$ is the original message token without the first class token and $Z^{s,t} = [x_{1}^{s,t}, x_{2}^{s,t},..., x_{n_{d}}^{s,t}]$; $n_{d}$ is the number of the original message token; $x_{cf}^{s, t}$ is the cross-frame event prompts; $\text{Att}(.)$ indicates the self-attention layer. Then, we extract the message tokens as the input of the next $(s+1)$\textit{-th} layer of the event encoder (See Fig.~\ref{Event Prompts Generation Branch} feature extraction block). Let the event encoder $E_{e}$ with $S$ layers, the above process iterates $S$ times. In the end, we extract the $CLS$ tokens from the token sequence and then project them to the $D^{'}$ dimension using the linear projection $P_{cls} \in \mathbb{R}^{D \times D^{'}}$. The event embeddings $f^{e}$ are ultimately produced by averaging the $CLS$ token among $T$ event frames.

\subsection{Text Encoder} 
\label{Section: Text Encoder}
\noindent\textbf{Hybrid Text Prompts} After extracting event embeddings based on the event encoder, we subsequently employ hybrid text prompts~\cite{yao2023visual} to enhance model generalization ability across diverse dataset: 1) the hand-crafted text prompt {\fontfamily{qcr}\selectfont‘A drafted image of a} [\textit{class}]{\fontfamily{qcr}\selectfont.'}, where [\textit{class}] represents the category name. Each word is encoded into a $D_{p}$-dimension word embeddings, forming the final text token $P_{h}$; 2) the learnable text prompt $P_{l} = [P_{1}, P_{2},..., P_{n_{l}}, P_{class}.]$, where $i=1,2,...,n_{l}$ is a random initialized parameter with $D_{p}$ dimensions; $n_{l}$ denotes the number of the learnable text prompts; $P_{class}$ represents the word embeddings of [\textit{class}] and $[.]$ means the concatenation operation. 

\noindent\textbf{Content Prompts} Besides, the content prompts are introduced to enhance the model performance. The content prompts are generated by a simple two-layer MLP. Specifically, MLP takes the event embeddings $f^{e}$ and image embeddings $f^{i}$ as input and produces the corresponding event content prompts $\sigma_{e}\in \mathbb{R}^{1 \times D_{p}} $ and the image content prompts $\sigma_{i}\in \mathbb{R}^{1 \times D_{p}}$, both of which have the same dimension $D_{p}$ as the hand-crafted text token $P_{h}$. Next, the learnable text prompts $P_{l}$ and the hand-crafted text prompts $P_{h}$ are pointwisely added with the content prompts $\sigma_{m}, m=i,e$, respectively, generating the final text input, as depicted in Fig.~\ref{pic:framework}. Next, the learnable text embeddings $f^{t,m}_{l}, m=i,e$ and hand-crafted text embeddings $f^{t,m}_{h}, m=i,e$ with content prompts are generated by the text encoder $E_t$. Finally, the text embeddings $f^{t,m}, m={e,i}$ of modality $m$ is obtained by averaging the $f^{t,m}_{l}$ and $f^{t,m}_{h}$.
\subsection{Hierarchical Triple Contrastive Alignment (HTCA)}\label{section: Hierarchical Triple Alignment}
With the extracted event embeddings $f^{e}$ (Sec.~\ref{Section: Domain-gap Shifted Event Encoder}), image embeddings $f^{i}$, and text embeddings with event content prompt $f^{t, e}$ (Sec.~\ref{Section: Text Encoder}), it's still an open question to bridge the modality gap among the three distinct modalities \ie, image, text, and event, and effectively align them. A simple objective is to align $f^{e}$ with $f^{t}$ from scratch. However, the relatively small event dataset and limited vocabulary may hinder reliable representation and plausible few-shot performance. To this end, we propose the Hierarchical Triple Contrastive Alignment (HTCA) module, aiming at aligning the event, image, and text modalities jointly and enhancing model performance based on semantic consistency.

\noindent \textbf{Multi-modality Triple Alignment} We first minimize the contrastive loss $L(f^{i}, f^{e})$ between the event embeddings $f^{e}$ and image embeddings $f^{i}$, as well as minimizing $L(f^{e}, f^{t,e})$ between the event embeddings $f^{t,e}$ and text embeddings with event content prompts $f^{t,e}$. We ignore the image-text pair because we empirically observe that altering the pre-trained image-text alignment reduces performance. Specifically, the contrastive loss is obtained by averaging the whole mini-batch among each pair of modalities $M_1$ and $M_2$, which can be formulated as follows: 
\begin{small}
\begin{align}
L(M^{1}, M^{2}) = \frac{1}{N} \sum_{n \in N}
-\log \frac{\exp \left(M^{1}_n \cdot M^{2}_n / \tau\right)}{\exp \left(M^{1}_n \cdot  M^{2}_n / \tau\right)+\sum_{m \in N, m \neq n} \exp \left(M^{1}_n \cdot M^{2}_m / \tau\right)},
\end{align}
\end{small}
where $\tau$ is the temperature coefficient, $N$ represents the size of the mini-batch, $n,m$ denote the $n$-th and the $m$-th data of the mini-batch, $m \neq n$ and $m = n$ indicates the negative pair and the positive pair in a batch respectively. 

\noindent \textbf{Semantic Consistency Alignment.}
Semantic consistency indicates that embeddings generated from different modalities of the same item share the same semantic meaning, \eg paired text and event data, resulting in the distance between them being closer in representation space. In light of this, we propose the semantic consistency alignment to enhance the model performance and inherit the remarkable few-shot ability of the pre-trained CLIP (Sec.~\ref{sec: Ablation Stduy}). Specifically, we ensure the semantic consistency between the image and event content added text embeddings $f^{t, i}$ and $f^{t, e}$ by employing loss function $L(f^{t, i}, f^{t, e})$, as well as additional $MSE( f^{t,m}_{l},  f^{t,m}_{h})$ loss to reduce the gap between text embeddings $f^{t,m}_{l}$ and $f^{t,m}_{h}$ generated by the learnable and hand-crafted prompts (Sec.~\ref{Section: Text Encoder}).

\subsection{Total Objective}
\label{Training Objectives}
In summary, as shown in Fig.~\ref{pic:framework}, the total objective is composed of $L(f^{i}, f^{e})$, $L(f^{e}, f^{t,e})$, as well as $L(f^{t, i}, f^{t, e})$ and $MSE( f^{t,m}_{l},  f^{t,m}_{h})$, all mentioned in Sec.~\ref{section: Hierarchical Triple Alignment}. We combine them with different trade-off hyper-parameters:
\begin{small}
\begin{align}
L_{final} = \alpha L(f^{i}, f^{e}) + \beta L(f^{e}, f^{t, e}) + \theta L(f^{t, i}, f^{t, e}) + \gamma MSE( f^{t,m}_{l},  f^{t,m}_{h}),
\end{align}
\end{small}

where $f^{e}, f^{i}, f^{t, e}, f^{t, i}$ denotes the event embeddings, image embeddings, event content added text embeddings and image content added text embeddings; $MSE(.)$ indicates the mean squared error loss function; $f^{t,m}_{l}$ and $f^{t,m}_{h}$ is the text embeddings generated from the learnable and hand-crafted text prompts with modality $m$ content prompts, respectively, where $m$ can be $e$ for event modality or $i$ for image modality. We set the default values of $\alpha$, $\beta$, $\theta$, and $\gamma$ to 1 as the frozen CLIP visual encoder serves as the initialization of the event encoder, ensuring the magnitudes of the four components distributed on the same scale. Note that when the image input is unavailable, only $L(f^{e}, f^{t,e})$ participates in backpropagation, namely setting $\alpha$, $\theta$, and $\gamma$ to 0 and $\beta$ to 1.

\section{Experiments and Evaluation}\label{sec:Experiments}
\subsection{Experimental Setup and Datasets}\label{sec:Datasets}

\textbf{Dataset.} We use three public image-event-text paired datasets for our experiments: N-ImageNet~\cite{kim2021n} with ImageNet-1K~\cite{deng2009imagenet}, N-Caltech101~\cite{fei2004learning} with Caltech101~\cite{orchard2015converting}, and N-MNIST~\cite{orchard2015converting} with MNIST~\cite{deng2012mnist}.
We split N-ImageNet and N-MNIST officially. N-Caltech101 is split 4:1 for training and validation. The same validation set is used for zero-shot, few-shot, and fine-tuning. Few-shot data are randomly selected and consistent. \textit{See more details on experimental settings in the Suppl.}

\noindent \textbf{Dataset Preprocessing.} RGB images are resized to $224\times224$ resolution to ensure consistency with ViT input settings, while grayscale images are transformed into a three-channel tensor by concatenating three identical copies. Then we transform the event data into event frames, with the aggregated event counts per frame $N$ being determined by hyperparameter search. \textit{See more discussion on $N$ in Sec.~\ref{sec: Ablation Stduy} and the details of event frame representation in suppl.}. 

\noindent \textbf{Inference and Task Settings.}
Event recognition indicates the classification of event data, which is a fundamental task in event-based vision. As shown in Fig.~\ref{TesearFig}, text class names are input into the text encoder to generate text embeddings $f^{t,e}$. The event data is encoded to produce the event embeddings $f^{e}$, and then the softmax function is applied to the final probability map generated by the multiplication between $f^{t,e}$ and $f^{e}$. In fine-tuning and few-shot event recognition scenarios, the former employs the entire training dataset, while the latter uses only a handful of training examples for each category during the training phase.

\subsection{The Results for Event Recognition Task}
\begin{table}[t!]
\centering
\caption{Fine-turned performance. ‘-' represents the exact number not provided.}
\setlength{\tabcolsep}{4mm}
\resizebox{1.0\textwidth}{!}{
\begin{tabular}{cccccc}
\hline
 &    &    & \multicolumn{3}{c}{Top-1 Accuracy (\%)}  \\ \cline{4-6} 
\multirow{-2}{*}{Method}  & \multirow{-2}{*}{Pretrain Dataset} & \multirow{-2}{*}{Labels}  & N-Caltech101 & N-MINIST & N-ImageNet \\ \hline
\rowcolor{gray!10}\multicolumn{6}{l}{\textit{Transfer learning of event self-supervised pre-training methods.}}    \\ \hline
HATS \cite{sironi2018hats}  & \usym{2717} & \usym{2717} & 64.20 & 99.10  & 47.14      \\
AEGNN \cite{schaefer2022aegnn}  & \usym{2717} & \usym{2717} & 66.80  & -  & -   \\
AsynNet \cite{messikommer2020event} & \usym{2717} & \usym{2717}   & 74.50 & -  & -   \\
EvS-S \cite{li2021graph} & \usym{2717} & \usym{2717} & 76.10  & -  & - \\
MEM\cite{klenk2024masked} & NImNet-1k & \usym{2717} & 90.10  & -  & 57.89 \\ \hline
\rowcolor{gray!10}\multicolumn{6}{l}{\textit{Transfer learning of image supervised pre-training methods.}} \\ \hline
RG-CNN\cite{bi2019graph} & \usym{2717} & \usym{1F5F8} & 65.70 & 99.00 & - \\
EST\cite{gehrig2019end} & ImNet-1k & \usym{1F5F8} & 81.70  & -  & 48.93 \\
DVS-ViT\cite{wang2022exploiting} & ImNet-21k  & \usym{1F5F8} & 83.00 & -  & -  \\
Matrix-LSTM\cite{cannici2020differentiable} & \usym{2717} & \usym{1F5F8} & 84.31 & 98.90 & 32.21 \\
E2VID\cite{rebecq2019events}  & ImNet-1k  & \usym{1F5F8} & 86.60  & 98.30  & - \\
DiST\cite{kim2021n} & \usym{2717} & \usym{1F5F8} & 86.81 & - & 48.43 \\
EventDrop\cite{gu2021eventdrop}  & ImNet-1k  & \usym{1F5F8} & 87.14  & -  & -  \\
ACE-BET\cite{liu2022fast} & ImNet-1k & \usym{1F5F8} & 89.95  & -  & -  \\
\hline
\rowcolor{gray!10}\multicolumn{6}{l}{\textit{Transfer learning of image-text supervised pre-training methods.}}\\ \hline
EventCLIP(ViT-L-14)\cite{wu2023eventclip} & WIT & \usym{1F5F8} & 93.57 & -  & 53.20 \\
\rowcolor{blue!5}EventBind(ours, ViT-B-32)  & WIT & \usym{1F5F8} & 93.74 & 99.26   & 42.94 \\ 
\rowcolor{blue!5}EventBind(ours, ViT-B-16)  & WIT & \usym{1F5F8} & 94.08 & 99.27   &51.40 \\
\rowcolor{blue!5}EventBind(ours, ViT-L-14)  & WIT & \usym{1F5F8} & \textbf{95.29}\textcolor{blue}{\fontsize{8}{8}\selectfont+1.72} & \textbf{99.45}\textcolor{blue}{\fontsize{8}{8}\selectfont+0.35}   & \textbf{63.54}\textcolor{blue}{\fontsize{8}{8}\selectfont+5.65} \\\hline
\end{tabular}}
\label{Fine-turned performance}
\end{table}

\noindent\textbf{Fine-tuning Setting.} Fine-tuning experiment results are presented in Tab.~\ref{Fine-turned performance} on N-Caltech101, N-MINIST, and N-ImageNet datasets compared with various baselines, including event self-supervised pre-training methods~\cite{sironi2018hats, schaefer2022aegnn, messikommer2020event, li2021graph, yang2023event, klenk2022masked, klenk2024masked} and transfer learning of supervised image pretraining methods~\cite{gehrig2019end, wang2022exploiting, rebecq2019events, gu2021eventdrop, liu2022fast}. \textit{our EventBind achieves new state-of-the-art accuracy compared with the previous method on three benchmarks by a large margin.} For N-Caltech101 dataset, our method achieves Top-1 accuracy of 95.29\%, surpassing the previous SOTA\cite{liu2022fast} by +5.34\% and competitor\cite{wu2023eventclip} by +1.72\%. EventBind achieves 63.54\% accuracy on N-ImageNet, an improvement of +4.79\% over SOTA \cite{klenk2024masked}. For N-MNIST, EventBind achieves SOTA accuracy of 99.45\%, compared to 99.10\% by~\cite{sironi2018hats}. 
The fine-tuning results prove EventBind's superiority in effectively aligning event data with text and image. Our EventBind adapts well to datasets with various object classes, including 10, 101, and 1,000, demonstrating its transferability and robustness. We employ three versions with ViT backbones: ViT-B-32, ViT-B-16, and ViT-L-14. \textit{See more discussion on model parameters with the different backbones in Sec.\ref{sec: Ablation Stduy}}.

\begin{table*}[t!]
\centering
\caption{The few-shot Top-1 accuracy (\%) compared with existing event-text model.}
\label{Zero-shot table}
\setlength{\tabcolsep}{5mm}
\resizebox{1.0\textwidth}{!}{
\begin{tabular}{cccccc}
\hline
& \multicolumn{3}{c}{EventBind (Ours)} & \multicolumn{2}{c}{EventCLIP\cite{wu2023eventclip}}\\ \cline{2-6} 
\multirow{-2}{*}{Setting} & N-Caltech101 & \multicolumn{1}{c}{N-Imagenet}  & N-MNIST  & \multicolumn{1}{c}{N-Caltech101} & \multicolumn{1}{c}{N-Imagenet} \\ \hline
5-shot  & \cellcolor{blue!5}\textbf{84.49}\textcolor{blue}{\fontsize{6}{6}\selectfont+0.92}  & \cellcolor{blue!5}\textbf{33.90}\textcolor{blue}{\fontsize{6}{6}\selectfont+2.78} & \cellcolor{blue!5}\textbf{94.44} & 83.57 &  31.12   \\
10-shot & \cellcolor{blue!5}\textbf{88.73}\textcolor{blue}{\fontsize{6}{6}\selectfont+1.31}  & \cellcolor{blue!5}\textbf{36.97}\textcolor{blue}{\fontsize{6}{6}\selectfont+2.73} & \cellcolor{blue!5}\textbf{95.76} & 87.42 &  34.24   \\
20-shot &  \cellcolor{blue!5}\textbf{92.11}\textcolor{blue}{\fontsize{6}{6}\selectfont+1.70} & \cellcolor{blue!5}\textbf{43.27}\textcolor{blue}{\fontsize{6}{6}\selectfont+3.99} & \cellcolor{blue!5}\textbf{97.43} & 90.41 &  38.28   \\ \hline 
\end{tabular}}
\end{table*}

\noindent\textbf{Few-shot Setting.}
\label{sec:Zero-shot and Few-shot Experiments}
We compare EventBind with existing methods, EventCLIP~\cite{wu2023eventclip}. To ensure comparison consistency, the ViT-L-14 backbone is utilized as EventCLIP stated. In Tab.~\ref{Zero-shot table}, EventBind beats EventCLIP in few-shot cases, with gains of 0.92\%, 1.31\% and 1.7\% for 5-shot, 10-shot, and 20-shot on N-Caltech101. Remarkably, our 20-shot accuracy outperforms the previous state-of-the-art (SOTA)~\cite{liu2022fast}, which was trained on the full dataset, while EventBind uses just 46.09\% of the data from the N-Caltech101 dataset. For the N-ImageNet dataset, EventBind also outperforms EventCLIP with 2.78\%, 2.73\%,
and 3.99\% accuracy improvements for 5-shot, 10-shot, and 20-shot settings, showcasing its superiority in low-shot scenarios.

\subsection{Ablation Study}\label{sec: Ablation Stduy}

\noindent \textbf{Effectiveness of EventBind’s
key components.}
\label{sec:The ablation study on overall network design}
Results in Tab.~\ref{AB_on_modules} demonstrate the HTCA module boosts performance by +42.14\% and +77.72\% on the N-Caltech and N-MNIst datasets. Text prompts further improve accuracy by +43.00\% and +80.18\%. Combining them with the proposed event encoder further increases accuracy to 93.87\% and 99.20\%, showing +43.68\% and +80.50\% gains, thus proving the effectiveness of EventBind’s key components.

\noindent\begin{minipage}[c]{0.49\textwidth}
		\centering
		\setlength{\tabcolsep}{0.6 mm}
  	    \captionof{table}{Ablation study on EventBind's key components.}
		\scalebox{0.60}{
\begin{tabular}{ccccc}
\hline
\multicolumn{3}{c}{Ablation Settings}  & \multicolumn{2}{c}{Top-1 Accuracy (\%)}  \\ \hline
HTCA & Text prompt & Event encoder & N-Caltech101 & \multicolumn{1}{c}{N-MNIST} \\ \hline
\usym{2717}  &\usym{2717} &  \usym{2717} & 50.40 & 18.77\\
\usym{1F5F8} &\usym{2717} &  \usym{2717} & 92.54\textcolor{blue}{\fontsize{8}{8}\selectfont+42.14}& 96.49\textcolor{blue}{\fontsize{8}{8}\selectfont+77.72}\\
\usym{1F5F8} & \usym{1F5F8}  &  \usym{2717} & 93.40\textcolor{blue}{\fontsize{8}{8}\selectfont+43.00} &  98.95\textcolor{blue}{\fontsize{8}{8}\selectfont+80.18}\\
\usym{1F5F8} & \usym{1F5F8} & \usym{1F5F8} & \textbf{94.08}\textcolor{blue}{\fontsize{8}{8}\selectfont+43.68}  & \textbf{99.27}\textcolor{blue}{\fontsize{8}{8}\selectfont+80.50}\\ \hline
\end{tabular}
\label{AB_on_modules}
	}
\end{minipage}
\begin{minipage}[c]{0.49\textwidth}
		\centering
		\setlength{\tabcolsep}{0.5mm}
            \captionof{table}{Ablation study on the HTCA module. `\_` highlights the EventBind performance in the absence of image inputs.}
		\scalebox{0.66}{
\begin{tabular}{ccccc}
\hline
\multicolumn{3}{c}{Ablation Settings} & \multicolumn{2}{c}{Top-1 Accuracy(\%)}\\ \hline
$L(f^{e}, f^{i})$ & $L(f^{e}, f^{t, e})$ & $L(f^{t, i}, f^{t, e})$ & N-Caltech101 & \multicolumn{1}{c}{N-MNIST} \\ \hline
\usym{2717}&\usym{2717}&\usym{2717}&50.40 & 18.77\\
\usym{1F5F8}&\usym{2717}&\usym{2717}&70.31\textcolor{blue}{\fontsize{8}{8}\selectfont+19.91} & 57.95\textcolor{blue}{\fontsize{8}{8}\selectfont+39.18} \\
\usym{2717}&\usym{1F5F8}&\usym{2717}&90.35\textcolor{blue}{\fontsize{8}{8}\selectfont+39.95} & 97.61\textcolor{blue}{\fontsize{8}{8}\selectfont+78.84}\\
\underline{\usym{1F5F8}}&\underline{\usym{1F5F8}}&\underline{\usym{2717}}&\underline{93.11}\textcolor{blue}{\fontsize{8}{8}\selectfont+42.71} &\underline{99.02}\textcolor{blue}{\fontsize{8}{8}\selectfont+80.25}\\
\usym{1F5F8}&\usym{1F5F8}&\usym{1F5F8}&\textbf{94.08}\textcolor{blue}{\fontsize{8}{8}\selectfont+43.68} &\textbf{99.27}\textcolor{blue}{\fontsize{8}{8}\selectfont+80.50}\\ \hline
\end{tabular}
\label{AB_tri_loss}}
\end{minipage}

\noindent
\begin{minipage}[c]{0.4\textwidth}
		\centering
		\setlength{\tabcolsep}{0.5 mm}
  	    \captionof{table}{Ablation study on event encoder. $P_{e}$ means event prompts.}
		\scalebox{0.61}{
\begin{tabular}{cccc}
\hline
\multicolumn{2}{c}{Setting}                  & \multicolumn{2}{c}{Top-1 Accuracy (\%)}   \\ \hline
$P_{e}$ & Event Temporal modeling & N-Caltech101 & N-MNIST \\ \hline
\usym{2717}  & \usym{2717}  & 90.40 & 95.78 \\
\usym{1F5F8} & \usym{2717}  & 91.85\textcolor{blue}{\fontsize{8}{8}\selectfont+1.45} &97.87\textcolor{blue}{\fontsize{8}{8}\selectfont+2.09}   \\
\usym{2717}  & \usym{1F5F8} & 92.27\textcolor{blue}{\fontsize{8}{8}\selectfont+1.87}&98.22\textcolor{blue}{\fontsize{8}{8}\selectfont+2.44} \\
\usym{1F5F8} & \usym{1F5F8} & \textbf{94.08}\textcolor{blue}{\fontsize{8}{8}\selectfont+3.68} &\textbf{99.27}\textcolor{blue}{\fontsize{8}{8}\selectfont+3.49}   \\ \hline
\end{tabular}
\label{AB_event_prompt}}
\end{minipage}
\begin{minipage}[c]{0.6\textwidth}
		\centering
		\setlength{\tabcolsep}{0.6mm}
            \captionof{table}{Model parameter analysis.}
		\scalebox{0.65}{
\begin{tabular}{cccc}
\hline
Backbone & ViT-L-14 & ViT-B-32 & ViT-B-16 \\ \hline
EventCLIP\cite{wu2023eventclip} & 890 MB & - & 335 MB \\
EventBind (w.o Image Encoder) & 1372 MB & 474 MB & 471 MB \\
EventBind (w.t. Image Encoder) & 1952 MB & 641 MB& 635 MB\\ \hline
\end{tabular}\label{ab_model_parameter}}
\end{minipage}

\noindent \textbf{The effectiveness of HTCA module}\label{sec:ab_study on HTCA module}
In Tab.~\ref{AB_tri_loss}, loss $L(f^{e}, f^{i})$ boosts accuracy by +19.91\% and 39.18\% on N-Caltech101 and N-MNIST. $L(f^{e}, f^{t,e})$ also enhances accuracy by +39.95\% and 78.84\%. EventBind combining $L(f^{e}, f^{i})$ and $L(f^{e}, f^{t,e})$ achieves +42.71\% and +80.25\%, demonstrating the effectiveness of EventBind when the image inputs are not available with slightly performance loss compared with the full settings(-0.97\%). With $L(f^{t,i}, f^{t,e})$, the whole HTCA module improves accuracy by +43.68\% and +80.50\%, showing the effectiveness of our proposed HTCA module.

\noindent \textbf{The effectiveness of event encoder.}
\label{sec:The ablation study on event encoder}
In Tab.~\ref{AB_event_prompt}, event prompts $P_{e}$ boost recognition accuracy by +1.45\% and +2.09\% on N-Caltech101 and N-MNIST datasets. Cross-frame prompts also enhance accuracy by +1.87\% and +2.44\%. Using both prompts together achieves +3.68\% and +3.49\% accuracy improvements. These results show the effectiveness of our event encoder.

\noindent \textbf{Model parameter size analysis.} We compare EventBind's parameter size across different ViT backbones with the current event-text model\cite{wu2023eventclip} in Tab.~\ref{ab_model_parameter}. Given that EventCLIP can solely generate text and event embeddings, EventBind incorporates an additional image encoder to link events, images, and text. For a fair comparison, we present the parameter size of EventBind with and without the image encoder. In Tab.~\ref{ab_model_parameter} and Tab.~\ref{Fine-turned performance}, EventBind (ViT-B-16) beats EventCLIP (ViT-L-14) on N-Caltech101 by 0.51\% using only 71.3\% of EventCLIP's parameters (635MB V.S. 890MB), showing its superior efficiency.

\begin{figure}[h]
    \centering
    \begin{subfigure}{0.32\textwidth} 
        \centering
        \includegraphics[width=\linewidth]{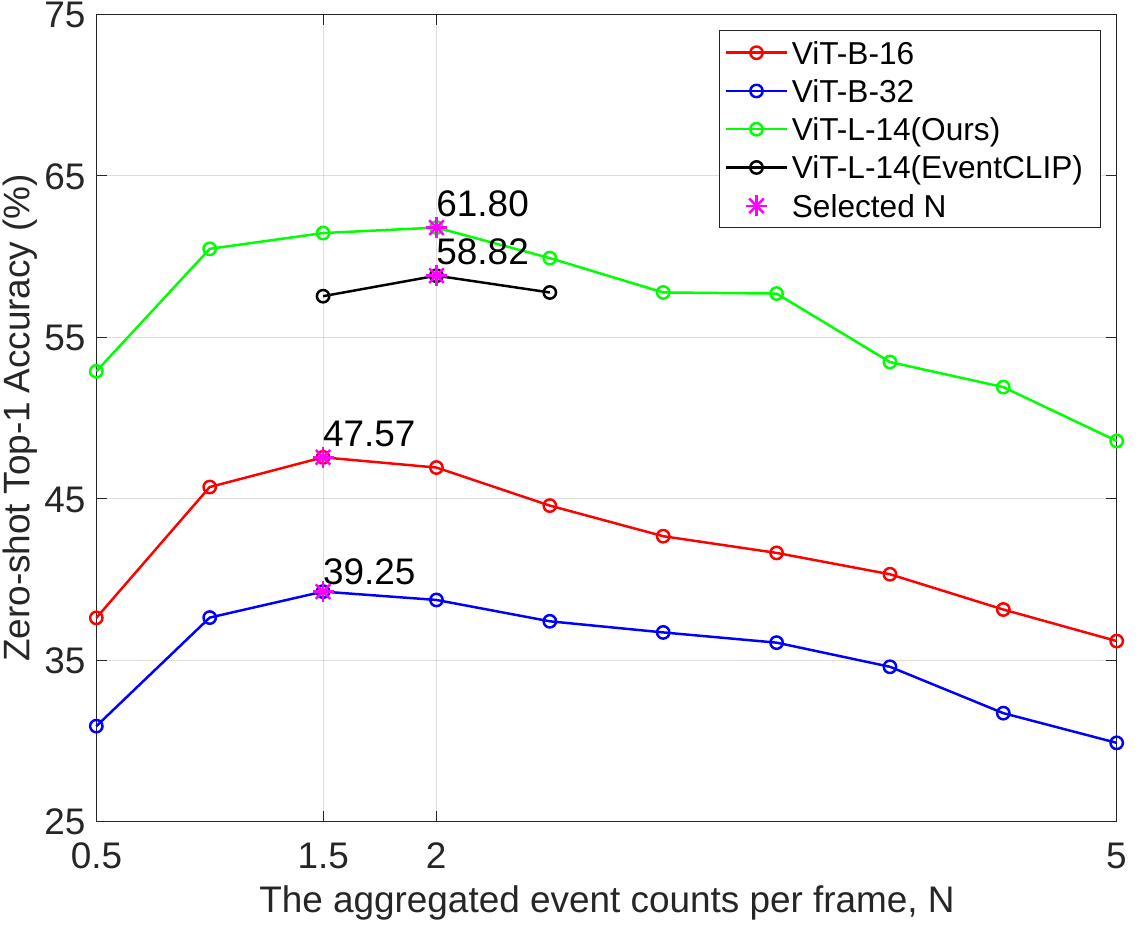}
        \caption{(a) N-Caltech101} 
        \label{fig:N-Caltech101}
    \end{subfigure}%
    \hspace{0.01\textwidth} 
    \begin{subfigure}{0.32\textwidth}
        \centering
        \includegraphics[width=\linewidth]{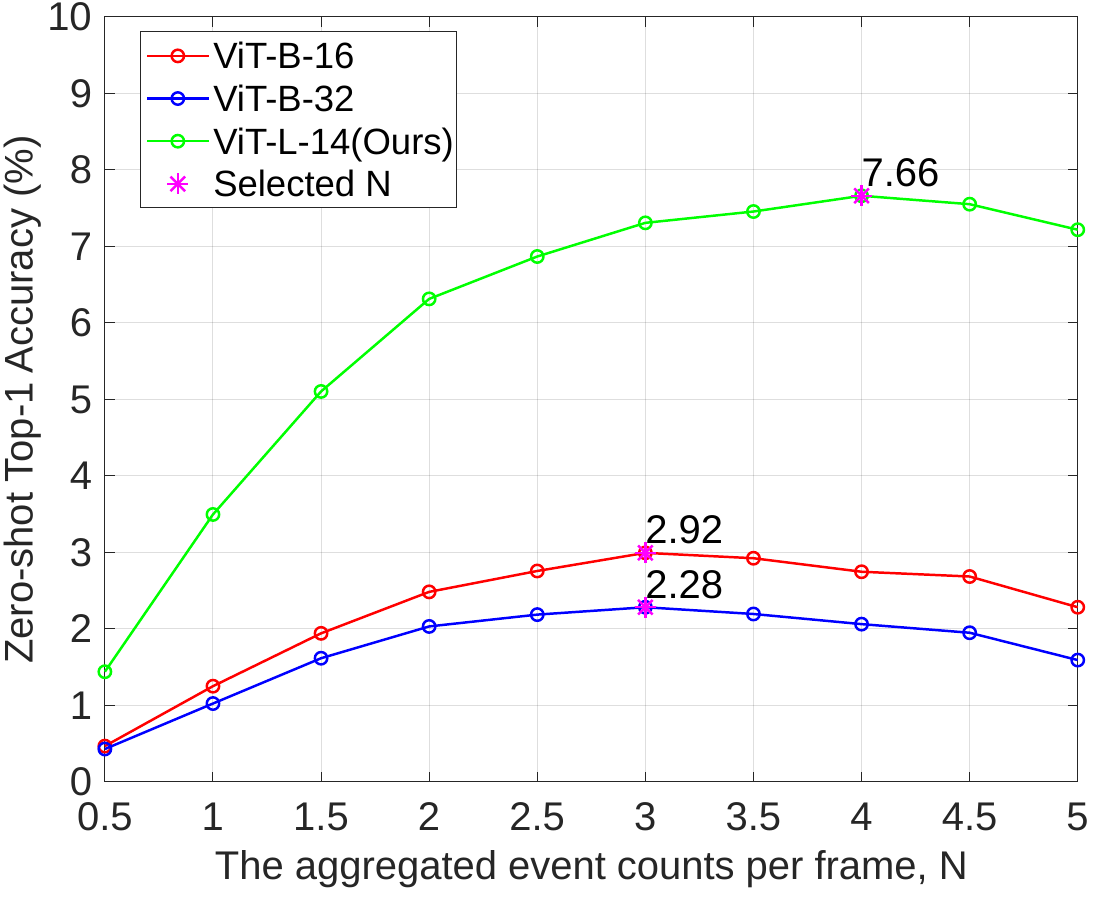}
        \caption{(b) N-Imagenet} 
        \label{fig:N-Imagenet}
    \end{subfigure}%
    \hspace{0.01\textwidth}
    \begin{subfigure}{0.32\textwidth}
        \centering
        \includegraphics[width=\linewidth]{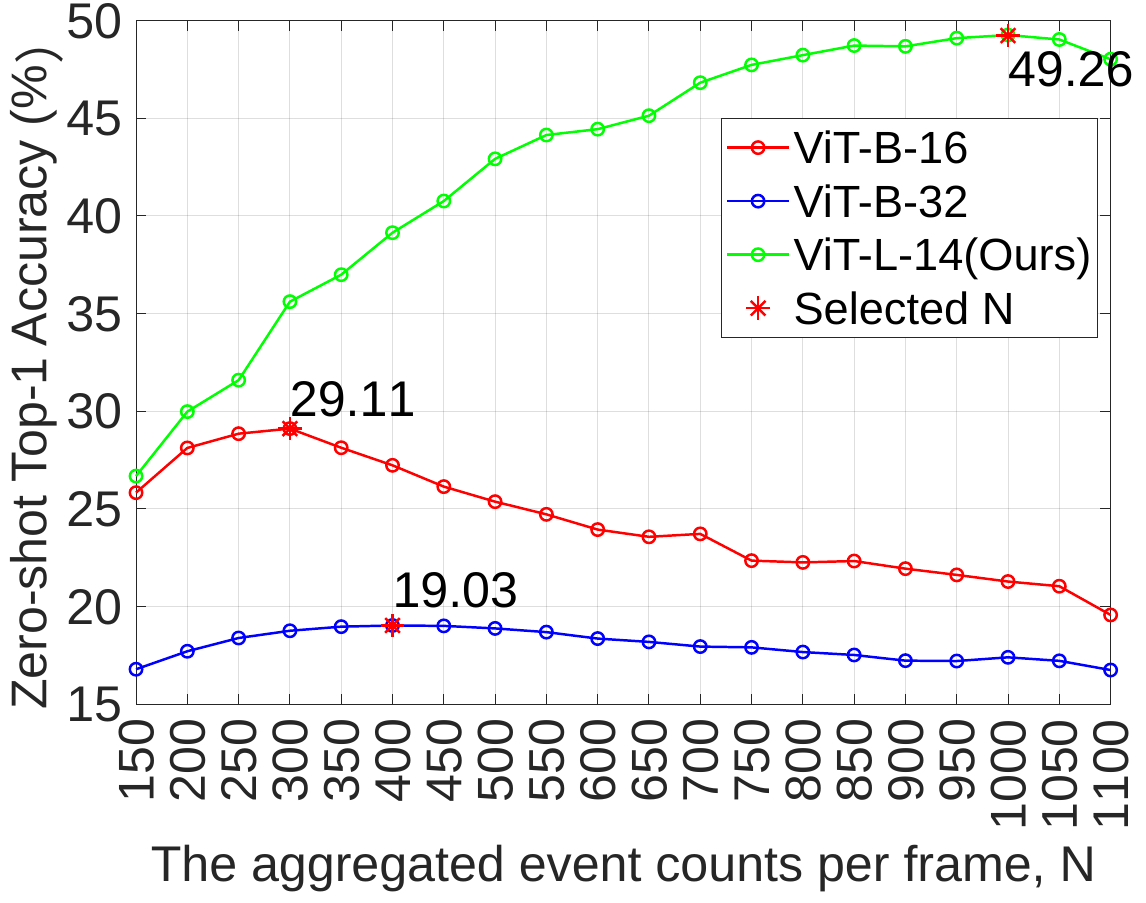} 
        \caption{(c) N-MNIST} 
        \label{fig:N-MNIST}
    \end{subfigure}
    \caption{Ablation of hyperparameters: the aggregated event counts per frame N based on zero-shot performance with 3 different ViT backbones across three datasets.}
    \label{fig:Hyper-parameter Search}
\end{figure}

\noindent \textbf{The aggregated event counts per frame $N$.} As shown in Fig.~\ref{fig:Hyper-parameter Search}, we set $N$ in training phrase based on the best zero-shot settings. We evaluate EventBind's zero-shot capability against EventCLIP on identical N-Caltech101 validation datasets, maintaining the reported hyperparameters but substituting the event-to-frames codes. EventCLIP achieves 58.82\% accuracy with 200,000 $N$, showing a -2.98\% decrease in zero-shot accuracy, highlighting EventBind's superiority.

\noindent \textbf{Event representation.}
\label{sec:Event representation}
Tab.~\ref{AB_training_configuration} displays the impact of different colorizing methods on frame-like event representations. Gray-scale achieves 92.18\% and 97.61\% accuracy, while RGB improves it to 94.08\% and 99.27\%.

\noindent\begin{minipage}[c]{0.4\textwidth}
		\centering
		\setlength{\tabcolsep}{0.5 mm}
  	    \captionof{table}{Ablation study on training configuration \& event frame representation.}
		\scalebox{0.75}{
\begin{tabular}{ccc}
\hline
\multirow{2}{*}{Setting} & \multicolumn{2}{c}{Top-1 Accuracy (\%)} \\ \cline{2-3} 
& N-Caltech101 & \multicolumn{1}{c}{N-MNIST} \\ \hline
Frozen CLIP  & 86.69  &  97.11  \\
Fitune CLIP  & \textbf{94.08}  & \textbf{99.27}  \\ \hline
Gray-scale & 92.18 & 97.61 \\
RGB & \textbf{94.08} & \textbf{99.27}  \\ \hline
\end{tabular}\label{AB_training_configuration}}
\end{minipage}
\begin{minipage}[c]{0.55\textwidth}
		\centering
		\setlength{\tabcolsep}{0.5mm}
            \captionof{table}{Ablation study on different hand-crafted text prompts.}
		\scalebox{0.7}{
\begin{tabular}{ccc}
\hline
\multirow{2}{*}{Hand-crafted text prompts} & \multicolumn{2}{c}{Top-1 Accuracy (\%)}   \\ \cline{2-3} 
& N-Caltech101 & N-MNIST \\ \hline
{\fontfamily{qcr}\selectfont‘A photo of a'} & 93.06 & 98.53 \\
{\fontfamily{qcr}\selectfont‘A sketch image of a'} & 93.89 & 99.20 \\
{\fontfamily{qcr}\selectfont‘A point cloud image of a'} & 92.71 & 98.49 \\
{\fontfamily{qcr}\selectfont‘An event frame of a'} & 92.41 & 98.87 \\ 
{\fontfamily{qcr}\selectfont‘A drafted image of a'} & \textbf{94.08} & \textbf{99.27} \\\hline
\end{tabular}\label{AB_Hand-crafted_text_prompts_prompt}}
\end{minipage}

\noindent\begin{minipage}[c]{0.4\textwidth}
		\centering
		\setlength{\tabcolsep}{0.5 mm}
  	    \captionof{table}{Abaltion study on the content prompts generation.}
		\scalebox{0.7}{
\begin{tabular}{cccc}
\hline
\multicolumn{2}{c}{Abaltion   settings} & \multicolumn{2}{c}{Top-1 Accuracy (\%)}           \\ \hline
$\sigma_{i}$ & $\sigma_{e}$ & N-Caltech101 & \multicolumn{1}{c}{N-MNIST}\\ \hline
\usym{2717} &\usym{2717} & 91.96 & 98.51\\
\usym{2717} &\usym{1F5F8} & 92.25\textcolor{blue}{\fontsize{8}{8}\selectfont+0.29}& 98.67\textcolor{blue}{\fontsize{8}{8}\selectfont+0.16}\\
\usym{1F5F8} &\usym{2717} & 92.82\textcolor{blue}{\fontsize{8}{8}\selectfont+0.86} & 98.72\textcolor{blue}{\fontsize{8}{8}\selectfont+0.21}\\
\usym{1F5F8} &\usym{1F5F8} & \textbf{94.08}\textcolor{blue}{\fontsize{8}{8}\selectfont+2.12}& \textbf{99.27}\textcolor{blue}{\fontsize{8}{8}\selectfont+0.76}\\  \hline
\end{tabular}\label{content prompts generation}}
\end{minipage}
\begin{minipage}[c]{0.275\textwidth}
		\centering
		\setlength{\tabcolsep}{0.5 mm}
  	    \captionof{table}{Ablation study on length of event modality prompts.}
		\scalebox{0.7}{
\begin{tabular}{ccc}
\hline
\multirow{2}{*}{$n_{e}$} & \multicolumn{2}{c}{Top1 Accuracy}  \\ \cline{2-3} 
& N-Caltech101 & N-MNIST \\ \hline
4    & 93.27  &  97.95  \\
8    & 93.89  &  98.73  \\
16   & \textbf{94.08}  &  \textbf{99.27}  \\ \hline
\end{tabular}\label{event modality prompts}}
\end{minipage}
\begin{minipage}[c]{0.275\textwidth}
		\centering
		\setlength{\tabcolsep}{0.5 mm}
  	    \captionof{table}{Ablation study on length of text learnable prompt.}
		\scalebox{0.7}{
\begin{tabular}{ccc}
\hline
\multirow{2}{*}{$n_{l}$} & \multicolumn{2}{c}{Top1 Accuracy} \\ \cline{2-3} 
& N-Caltech101 & N-MNIST \\ \hline
4   & 93.33 & 98.10 \\
8   & 93.75 & 98.83 \\
16  & \textbf{94.08} & \textbf{99.27} \\ \hline
\end{tabular}\label{text learnable prompt}}
\end{minipage}

\noindent \textbf{Fine-tune CLIP vs. Frozen CLIP as the event encoder backbone.}
\label{sec:Fine-tune CLIP V.S. Frozen CLIP as Event-Encoder backbone}
In Tab.~\ref{AB_training_configuration}, we ablate to freeze and fine-tune the CLIP image encoder of Event Encoder. Using the frozen CLIP image encoder yields 86.69\% and 97.11\% accuracy while fine-tuning it raises accuracy to 94.08\% and 99.27\%, respectively.

\noindent \textbf{The initialization of hand-crafted text prompts.}
We ablate five unique text prompts to determine which text embeddings best align with the event embeddings. The prompt {\fontfamily{qcr}\selectfont‘A drafted image of a} [\textit{class}]{\fontfamily{qcr}\selectfont.'} presents the best results of 94.08\% and 99.27\% on both N-Caltech101 and N-MNIST datasets.

\noindent \textbf{The event and image content prompts.} In Tab.~\ref{content prompts generation}, 
using the image content prompts $\sigma_{i}$ alone boosts accuracy by +0.29\% and +0.16\% while employing the event content prompts $\sigma_{e}$ alone improve accuracy by +0.86\% and +0.21\% on N-Caltech101 and N-MNIST datasets. Combining both yields a higher accuracy gain of +2.12\% and +0.76\%. Results show the effectiveness of our content prompt generation in boosting performance.

\noindent \textbf{The length of text learnable prompts $n_{l}$.}
In Tab.~\ref{event modality prompts}, we evaluate different lengths of learnable text prompts, i.e., 4, 8, and 16. Using a length of 16 for the event modality prompts results in the highest recognition accuracy.

\noindent \textbf{The length of event modality prompts $n_{e}$.} In Tab.~\ref{text learnable prompt}, We evaluate event modality prompts of varying lengths, namely 4, 8, and 16. Employing 16 as the prompt length yields the highest recognition accuracy.

\subsection{Extension Experiment: Event Retrieval}

\begin{figure}[t!]
\centering
\includegraphics[width=\linewidth]{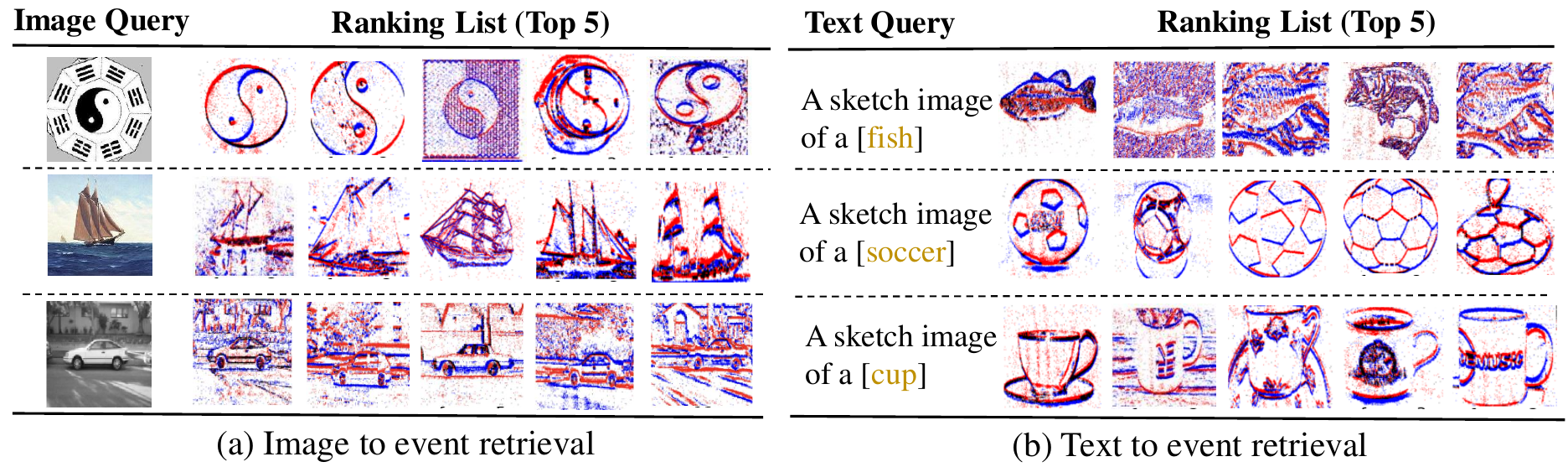}
\caption{Visual examples for the event retrieval task.}\label{EventsRetrieval}
\end{figure}
EventBind can be expanded to image-to-event and text-to-event retrieval tasks. \textit{See more details about event retrieval and results discussions in Suppl.} We initially perform the event
retrieval task by employing both text and image queries. In Fig.~\ref{EventsRetrieval}, we present the Top-5 event data retrieved from the N-Caltech101 dataset using randomly selected images or class texts as queries. All the retrieved event data exhibit a very high degree of similarity to the input text query, showcasing the effectiveness of our proposed EventBind framework.

\section{Conclusion and Future Work}
In this paper, we tackled the challenge of generalizing CLIP to event modality, thus establishing a unified representation space among images, text, and events. To this end, we proposed EventBind, a novel framework that capitalizes on the potential of CLIP for event-based recognition tasks to compensate for the lack of large-scale event-based datasets. We introduced novel event encoders and text prompts designed to exploit the unique properties of event data and enhance its generalization ability. Furthermore, we proposed a hierarchical triple contrastive alignment to optimize the correlation alignment among these three distinct modalities. Extensive experiments on three object recognition benchmarks demonstrated the superiority and efficiency of EventBind. Extension experiments on event retrieval indicated it possesses excellent generalization ability. 

\noindent\textbf{Future work}. We plan to extend EventBind to event-based downstream tasks. We hope that EventBind will facilitate the incorporation of knowledge from the text modality into the event-based vision, thereby harnessing decades of valuable research and contributing to the development of a more principled framework.

\section*{Acknowledgements}
This paper is supported by the National Natural Science Foundation of China (NSF) under Grant No. NSFC22FYT45, the Guangzhou City, University and Enterprise Joint Fund under Grant No.SL2022A03J01278, and Guangzhou Fundamental and Applied Basic Research (Grant Number: 2024A04J4072)

\section*{Appendix}
\section*{A. Preliminary of CLIP}
The Contrastive Language-Image Pre-training (CLIP)~\cite{CLIP} model is a prominent vision-language model that combines an image encoder (ViT\cite{dosovitskiy2020image} or ResNet\cite{he2016deep}), with a text encoder based on the Transformer~\cite{vaswani2017attention} architecture. CLIP focuses on generating visual and text embeddings and employs a contrastive loss to align these embeddings in a unified feature space. Notably, CLIP's exceptional transferability is attributed to its pretraining on a large-scale dataset of more than four million image-text pairs.

For CLIP-based recognition, it usually employs a simple yet efficient method that takes \emph{N} hand-crafted prompts like `a photo of a $[cls]$' as the input for the text encoder to obtain the \emph{D}-dimensional textual embedding $f^{t} \in \mathbb{R}^{N \times D}$, where $[cls]$ is the class name and \emph{N} is the number of classes of the downstream dataset. Meanwhile, for a given image, the visual embedding $f^{i} \in \mathbb{R}^{1 \times D}$ is obtained by the visual encoder. Then, the recognition logits $p \in \mathbb{R}^{N}$ can be obtained by calculating the similarity of text and visual embeddings by multiplication:
\begin{align}
    p = \mathrm{Softmax}(f^{i}(f^{t})^{T}),
\end{align}
where $\mathrm{Softmax}(.)$ denotes the Softmax function and \emph{p} represents the predicted probability of the \emph{N} classes. Finally, the highest scores of the logits \emph{p} is regarded as the final prediction \emph{P}:
    \begin{align}
    P = \arg\max_{i\in\mathcal N} p_{i}
\end{align}

\section*{B. Implementation Details}
\noindent \textbf{Event Frame-like Representation}
Event cameras capture object movement by detecting temporal independence and recording pixel-level brightness changes. The event stream, denoted as $\mathcal{E} = {e_{i}(x_i,y_i,t_i,p_i)}$, reflects the brightness change $e_{i}$ of a pixel at the timestamp $t_i$, with coordinates $(x_i,y_i)$, and polarity $p_i \in {1,-1}$. Here, 1 and -1 represent the positive and negative intensity change of brightness, respectively.
We convert the event stream into a sequence of frames. First, we normalize the length of the event stream $\mathcal{E}$ to a fixed amount $P$ using zero-padding or taking the first $P$ events. Then, we group every $Q$ consecutive event data in the normalized event stream $\mathcal{E}^{'}$ to obtain event parts $\mathcal{E}^{''} \in \mathbb{R}^{T \times 4 \times P}$, where $T = P/Q$ and denotes the number of event frames. 

We then transform these event parts into histograms $h \in \mathbb{R}^{T \times H \times W \times 2}$ by counting the amount of positive and negative events per pixel, where $H$ and $W$ represent the height and width of an event frame, respectively. Finally, to attain the grid-like 3-channel input $I_{e} \in \mathbb{R}^{T \times H \times W \times 3}$, we colorize the $t$th event histogram $h_{t}$ by multiplying the predefined red color map $[0,255,255]$ and blue color map $[255,255,0]$ with the positive and negative event histogram respectively and merge them by pixel-wise addition, which can be formulated as follow:

\begin{align}
    I_{e, t} = h_{t}^{p}[0,255,255]^{T} + h_{t}^{n}[255,255,0]^{T},
\end{align}
where the $I_{e, t}$ represents the event input for the $t$th event frame and $t=0,1,...,T$; $h_{t}^{p}$ and $h_{t}^{n}$ denotes the positive and negative event histogram respectively. Finally, the generated event frame input is resized into the 224 $\times$ 224 resolution for adapting to the ViT setting, generating the final event input $I_{e}^{'} \in \mathbb{R}^{T \times 224 \times 224 \times 3}$.

The above grid-like event representation is simple yet effective since the generated event image resembles the edge image, whose data distribution is much closer to the pre-trained natural images so that the transfer pressure is erased and the modality gap is bridged.

\begin{table*}[t!]
\centering
\begin{tabular}{ccccc}
\hline
Backbone & N-Imagnet & N-MNIST & N-Caltech101 \\ \hline
ViT-B-16 & 300,000 & 300 & 150,000  \\
ViT-B-32 & 300,000 & 400 & 150,000  \\
ViT-L-14 & 400,000 & 1,000 & 200,000 \\ \hline
\end{tabular}
\caption{The settings of aggregated event counts per frame $N$ for different backbones on three datasets.}
\label{N}
\end{table*}

\noindent \textbf{Additional Dataset Settings}
\textbf{N-ImageNet}~\cite{kim2021n} is derived from ImageNet-1K dataset, where the RGB images are displayed on a monitor and captured by a moving event camera. It includes 1,781,167 event streams with 480 × 640 resolution across 1,000 unique object classes. \textbf{N-Caltech101}~\cite{fei2004learning} contains event streams captured by an event camera in front of a mobile 180 × 240 ATIS system\cite{posch2010qvga} with the LCD monitor presenting the original RGB images in Caltech101. There are 8,246 samples comprising 300 ms in length, covering 101 different types of items. \textbf{N-MNIST} is created by displaying a moving image from the MNIST dataset on the ATIS system with the LCD monitor. It contains 70,000 event data samples covering 10 handwritten numbers from 0 to 9. 

\noindent \textbf{Additional Experimental Settings}
We use "A drafted image of a [\textit{CLS}]" as the hand-crafted text prompts template. The Pytorch~\cite{paszke2019pytorch} framework serves as the foundation for all experiments. The initial learning rates are set to 1e-5 for the N-Caltech101 dataset and 1e-6 for the N-Imagenet and N-MNIST datasets. The weight decay is 2e-4. CosineAnnealingLR~\cite{loshchilov2016sgdr} learning rate schedule is utilized, and the minimal learning rate is 1e-8. All few-shot and fine-tuning experiments are trained for 30 epochs with Adam~\cite{kingma2014adam} optimizer. Unless specified otherwise, the ablation study is conducted on the N-Caltech101/Caltech101 datasets utilizing ViT-B-16~\cite{dosovitskiy2010image} image encoder and the Transformer-based text encoder~\cite{vaswani2017attention} as the backbone. The event encoder is initialized with the CLIP image encoder's pre-trained weights and is fine-tuned during training.
We choose the aggregated event counts per frame $N$ based on the best EventBind's zero-shot performance. Tab.~\ref{N} represents the value of $N$ set for different backbones on three datasets.

\begin{table*}[t!]
\centering
\begin{tabular}{ccclcl}
\hline
\multicolumn{2}{c}{Ablation settings} & \multicolumn{2}{c}{Base}   & \multicolumn{2}{c}{New}    \\ \hline
Learnable & Hand-crafted & N-Caltech101 & \multicolumn{1}{c}{N-MNIST} & N-Caltech101 & \multicolumn{1}{c}{N-MNIST} \\ \hline
\usym{2717} & \usym{1F5F8} & 90.57 & 91.03 & \textbf{72.55}  & 18.80  \\
\usym{1F5F8} & \usym{2717} & 91.01 & 93.22 & 61.85  & 46.02  \\
\usym{1F5F8} & \usym{1F5F8} & \textbf{91.23} & \textbf{93.56} & 68.02  & \textbf{49.92} \\ \hline
\end{tabular}
\caption{Ablation study on hybrid text prompts module.}
\label{hybrid text prompts}
\end{table*}

\section*{C. Additional Experiment Result}
\noindent \textbf{The Effectiveness of Hybrid Text Prompts}
We ablated the hybrid text prompts to evaluate their impact on fine-tuning performance and generalization ability. The experiment follows the setting proposed in~\cite{zhou2022conditional}, where the entire dataset is divided into the base and new datasets, each containing half of the object classes. For each ablation setting, the model is fine-tuned on the base set without being exposed to half of the event classes in the new set. The evaluation is then performed on the new set to assess the generalization ability.

As shown in Tab.~\ref{hybrid text prompts}, the hand-crafted text prompts achieve an impressive 72.55\% Top-1 accuracy on the new set for the N-Caltech101 dataset, showcasing its remarkable zero-shot ability. In contrast, while learnable text prompts boost fine-tuning performance on the base set, they lead to decreased accuracy on the new set due to limited generalization capabilities. \textit{In contrast, our proposed hybrid text prompts module, which integrates both learnable and manually-crafted text prompts, attains the highest fine-tuning accuracy of 91.23\% while exhibiting a smaller decrease in zero-shot performance relative to solely utilizing single hand-crafted text prompts.} (\textbf{-10.70\% V.S -4.53\%} for N-Caltech101). For the N-MNIST dataset, it's expected that the model exhibits low accuracy on the base dataset with hand-crafted text prompts. This stems from the subpar image recognition capability of the original CLIP on N-MNIST, where zero-shot CLIP results are 10\% inferior to those of the linear probe on ResNet50 for MNIST.~\cite{CLIP}). \textit{Our model, equipped with the hybrid text prompts module, secures top-one accuracy on both the base dataset (93.56\%) and the new dataset (98.86\%) for N-MNIST, underscoring the effectiveness of our proposed module.}

\begin{table*}[t!]
\centering
\begin{tabular}{c|cc|cc}
\hline
& \multicolumn{2}{c|}{EventBind (scratch)} & \multicolumn{2}{c}{EventCLIP (scratch)} \\ \cline{2-5} 
\multirow{-2}{*}{Setting} & N-Calthech101 & N-MINIST & N-Calthech101 & N-MINIST \\ \hline
0-shot & \textbf{61.80} & \textbf{56.81} & 58.82 & 48.72 \\
1-shot & \textbf{74.96} & \textbf{74.64} & 70.53 & 74.62 \\
2-shot & \textbf{79.78} & \textbf{82.89} & 77.71 & 82.78 \\ \hline
\end{tabular}
\caption{The zero-shot and few-shot results.}
\label{Zero-shot}
\end{table*}

\begin{table}[t!]
\centering
\setlength{\tabcolsep}{1mm}
\resizebox{\linewidth}{!}{
\begin{tabular}{cccccc}
\hline
\multicolumn{1}{c|}{Model} & E2VID\cite{rebecq2019events} & SSL-E2VID\cite{paredes2021back} & Wang et al.\cite{wang2019event} & Ev-LaFOR\cite{cho2023label} & \textbf{EventBind(Ours)} \\ \hline
\multicolumn{1}{c|}{Top1 Accuracy} & 61.7 & 25.3 & 48.5 & 85.56 & \textbf{86.02} \\ \hline
\end{tabular}}
\caption{Open-vocabulary results on N-Caltech101.}
\label{open_world}
\end{table}

\noindent\textbf{Zero-shot, 1-shot, 2-shot Results Compared with EventCLIP.}
Since the dataset split of EventCLIP for 1-shot and 2-shot is \underline{unavailable}, we evaluated it on our dataset split with ViT-L-14 backbones. As shown in Tab.~\ref{Zero-shot}, our EventBind achieves better zero-shot, 1-shot, and 2-shot performance than EventCLIP. 

\noindent \textbf{Open-vocabulary Recognition Result}
For open-vocabulary event recognition, the text classes are arbitrary rather than restricted to the training classes. We follow the dataset split based on its open-source repository. The results are in Tab.~\ref{open_world}. Our EventBind achieves the SoTA performance of \textbf{86.02\%}, proving its superiority for open-vocabulary recognition.

\begin{figure}[t!]
    \centering
    \begin{subfigure}{0.32\textwidth} 
        \centering
        \includegraphics[width=\linewidth]{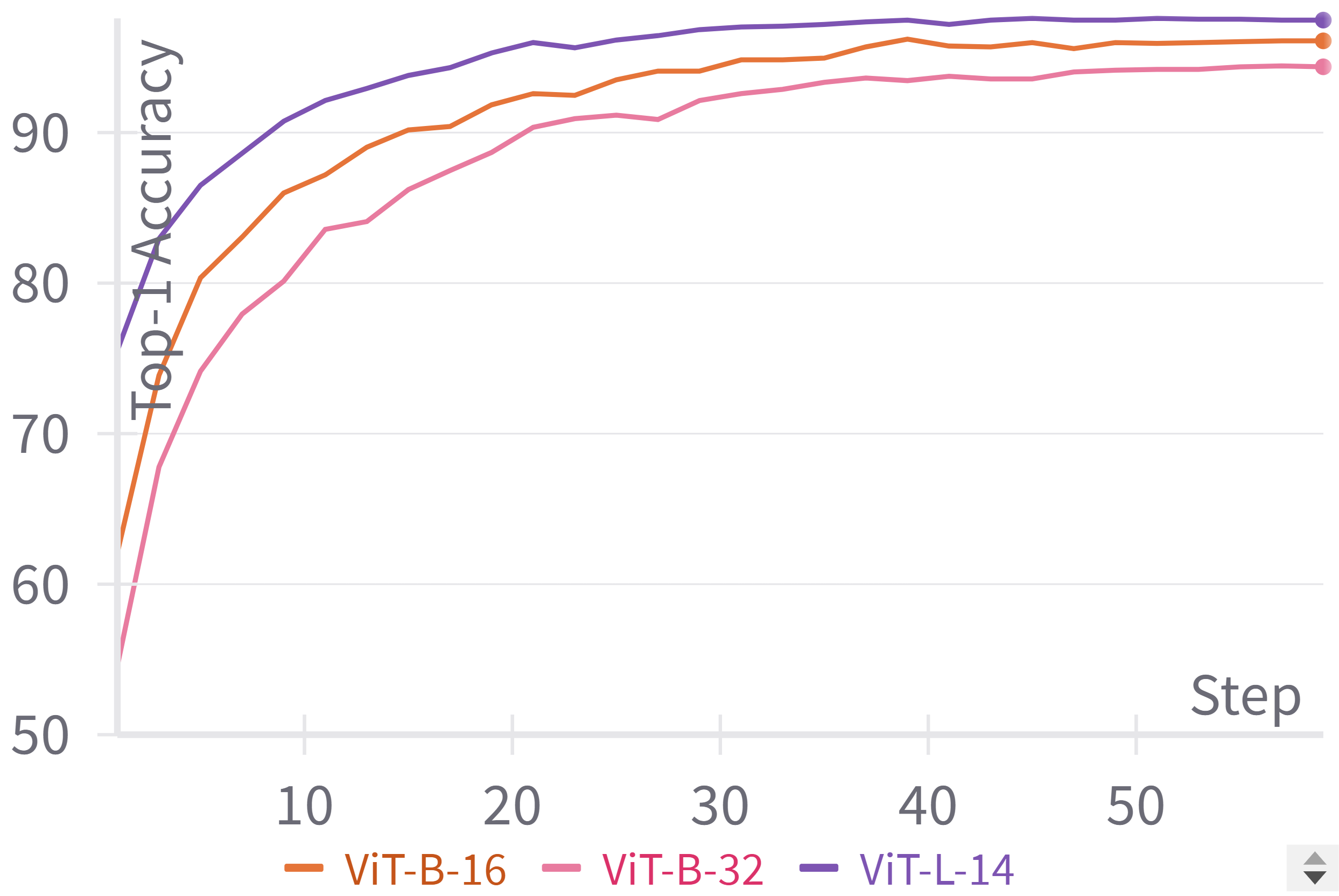}
        \caption{(a) N-Caltech101} 
        \label{fig:N-Caltech101}
    \end{subfigure}%
    \hspace{0.01\textwidth} 
    \begin{subfigure}{0.32\textwidth}
        \centering
        \includegraphics[width=\linewidth]{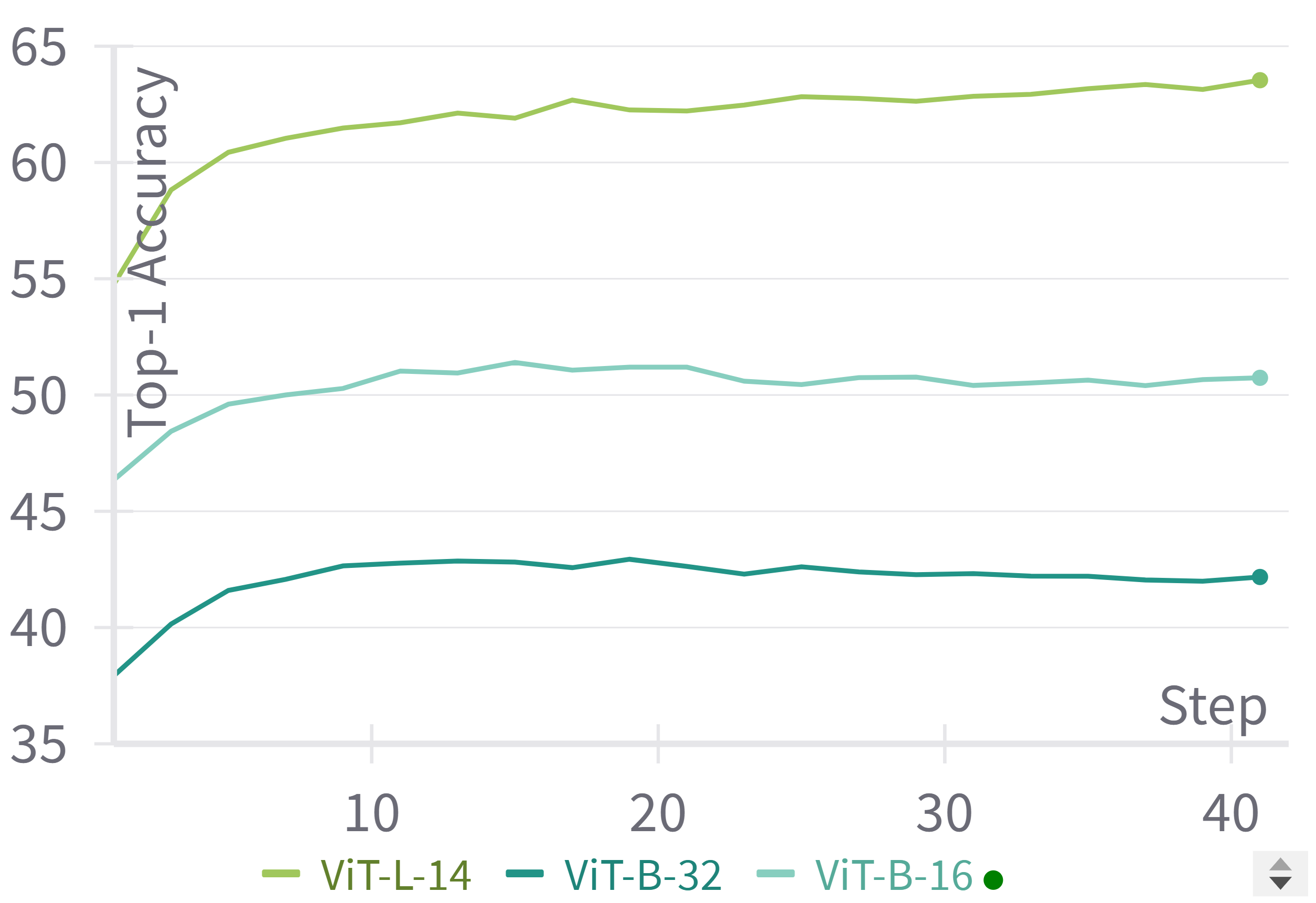}
        \caption{(b) N-Imagenet} 
        \label{fig:N-Imagenet}
    \end{subfigure}%
    \hspace{0.01\textwidth}
    \begin{subfigure}{0.32\textwidth}
        \centering
        \includegraphics[width=\linewidth]{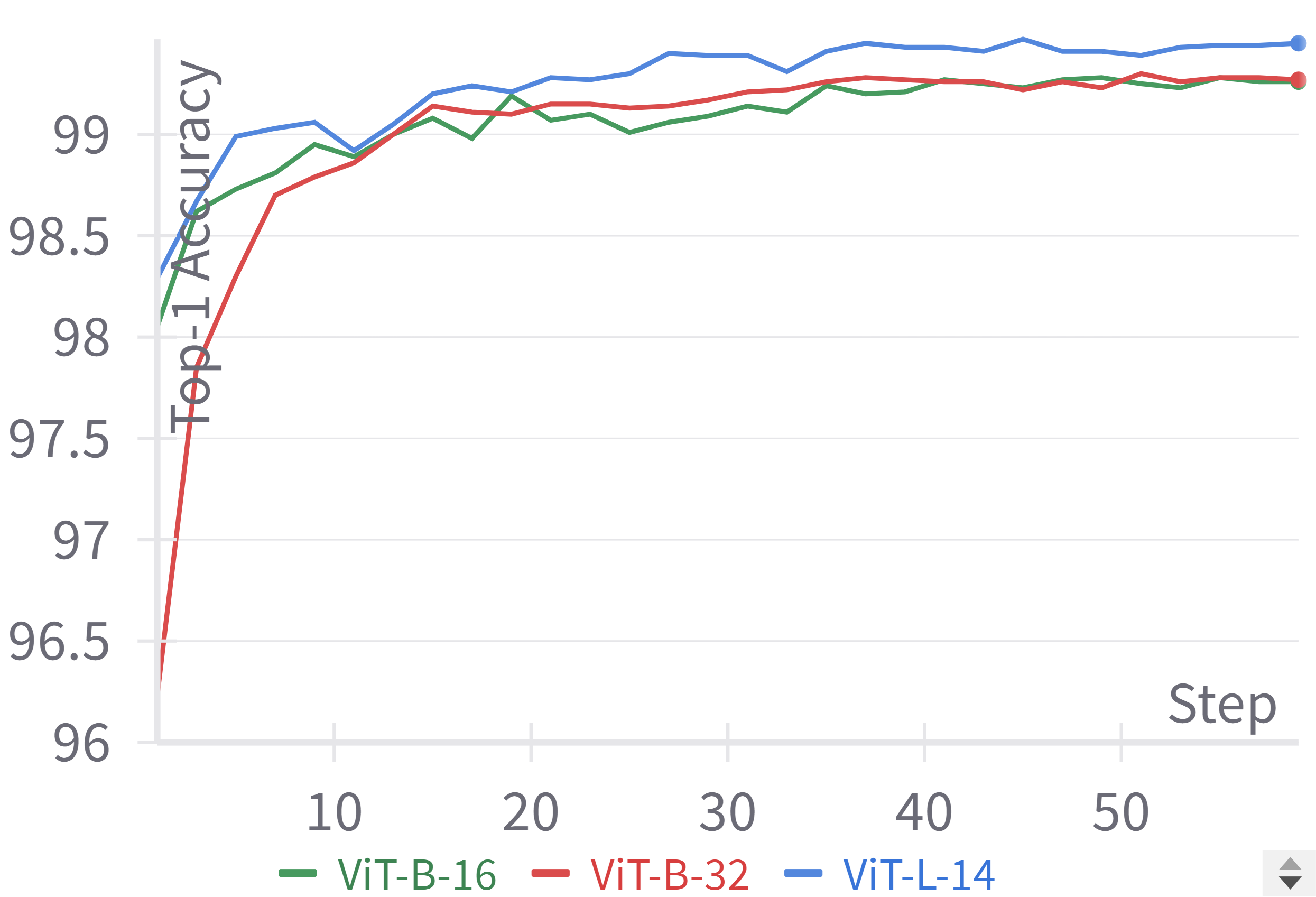} 
        \caption{(c) N-MNIST} 
        \label{fig:N-MNIST}
    \end{subfigure}    \caption{The Accuracy curves with three ViT backbones on the N-MINIST, N-Caltech101, and N-Imagenet datasets.}
    \label{Accuracy_curve_Ncaltech101}
\end{figure}

\noindent \textbf{Accuracy Curve}
We present the training accuracy curves trained by employing three ViT backbones on the N-MINIST, N-Caltech101, and N-Imagenet datasets in Fig.~\ref{Accuracy_curve_Ncaltech101}. As the number of training epochs increases, the accuracy of the model increases steadily, demonstrating the stability and reproducibility of EventBind.

\begin{table*}[t!]
\centering
\begin{tabular}{ccccccc}
\hline
& \multicolumn{3}{c}{Text2Event} & \multicolumn{3}{c}{Imgae2Event} \\ \cline{2-7} 
\multirow{-2}{*}{Setting} & Recall@1 & Recall@5 & Recall@10 & Recall@1 & Recall@5 & Recall@10 \\ \hline
0-shot & 78.22 & 88.12 & 91.09 & 79.40 & 90.34 & 93.58  \\ \hline
1-shot & 81.19 & 95.05 & 98.02 & 82.75 & 96.04 & 99.01  \\
2-shot & 90.10 & 99.01 & 100.00 & 88.31 & 98.02 & 99.01   \\
5-shot & 97.03 & 100.00 & 100.00 & 90.97 & 99.01 & 100.00 \\
10-shot & 98.02 & 100.00 & 100.00 & 94.73 & 99.01 & 100.00  \\
20-shot & 99.01 & 100.00 & 100.00 & 96.70 & 100.00 & 100.00\\\hline
Fine-tune & 100.00 & 100.00 & 100.00 & 96.88 & 99.60 & 99.95\\ \hline
\end{tabular}
\caption{Event Retrieval Results on N-Caltech101 Dataset with Zero-shot, Few-shot, and Fine-tuning Settings.}
\label{Event Retrieval label}
\end{table*}

\begin{figure}[t]
    \centering
    \begin{subfigure}{0.34\textwidth}
        \centering
        \includegraphics[width=\linewidth]{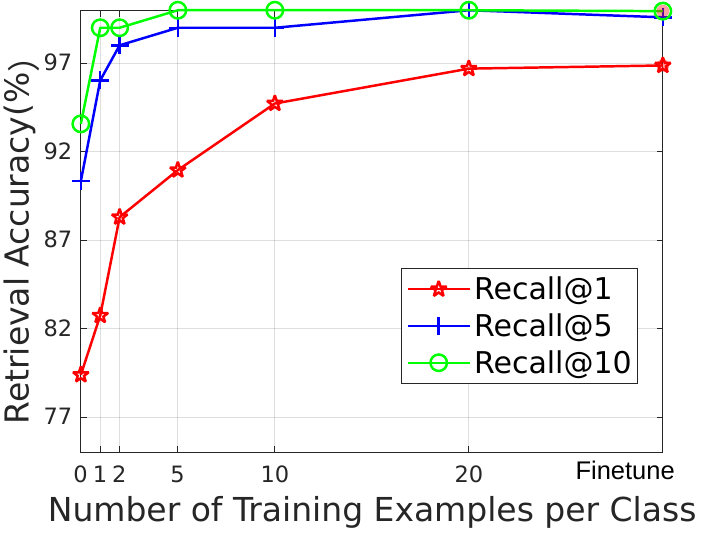}
        \caption{(a) Image-to-event retrieval.} 
        \label{fig:Image-to-event retrieval}
    \end{subfigure}%
    \hspace{0.01\textwidth}
    \begin{subfigure}{0.36\textwidth}
        \centering
        \includegraphics[width=\linewidth]{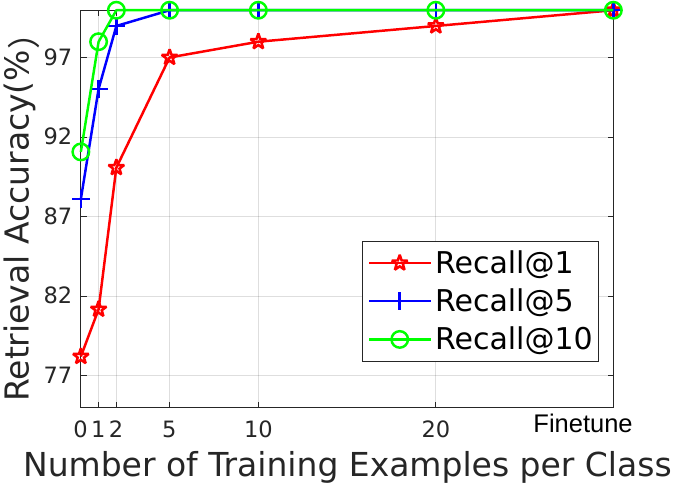} 
        \caption{(b) Text-to-event retrieval.} 
        \label{fig:Text-to-event retrieval}
    \end{subfigure}
    \caption{Event retrieval results on N-Caltech101 dataset with zero-shot, few-shot, and fine-tuning settings.}  
    \label{fig:Retrieval}
\end{figure}

\noindent \textbf{Event Retrieval Numerical Results}
We utilize Recall@1, Recall@5, and Recall@10 metrics commonly employed in retrieval tasks~\cite{young2014image,chen2015microsoft}. In Fig.~\ref{fig:Retrieval}, our EventBind shows remarkable retrieval performance (99.01\% Recall@1 for text query and 96.70\% Recall@1 for image query) with only 20 shot training examples, demonstrating its remarkable capabilities in few-shot learning. Notably, our model excels in text-to-event and image-to-event retrieval, achieving recall rates close to 100.00\% on Recall@1 after fine-tuning. This exceptional performance demonstrates that EventBind effectively establishes a unified representation space with aligned event, images and text embeddings. To ease future comparison, we report all those numbers in Tab. \ref{Event Retrieval label}.

%
%
\bibliographystyle{splncs04}
\bibliography{main}
\end{document}